\documentclass[10pt,twocolumn,letterpaper]{article}

\usepackage[pagenumbers]{wacv}

\definecolor{wacvblue}{rgb}{0.21,0.49,0.74}
\usepackage[pagebackref,breaklinks,colorlinks,allcolors=wacvblue]{hyperref}
\usepackage{tabularx}
\usepackage{url}
\usepackage{booktabs}
\usepackage{amsfonts}
\usepackage{nicefrac}
\usepackage{microtype}
\usepackage{graphicx}
\usepackage{algorithm}
\usepackage{amsmath}
\usepackage{algpseudocode}

\newcommand{\wacvacceptedwatermark}{%
  \AddToShipoutPicture*{%
    \AtPageUpperLeft{%
      \raisebox{-4.5mm}[0pt][0pt]{%
        \makebox[\paperwidth][c]{%
          \color[gray]{0.6}
           Accepted at the IEEE/CVF Winter Conference on Applications of Computer Vision (WACV) 2026%
        }%
      }%
    }%
  }%
}

\def\wacvPaperID{2153}
\def\confName{WACV}
\def\confYear{2026}

\title{DiRe: Diversity-promoting Regularization for Dataset Condensation}

\author{
Saumyaranjan Mohanty$^{1}$ \quad Aravind Reddy$^{2}$ \quad Konda Reddy Mopuri$^{1}$\\
{\small $^{1}$Department of Artificial Intelligence, Indian Institute of Technology Hyderabad}\\
{\small $^{2}$Centre for Responsible AI, Wadhwani School of Data Science \& AI, Indian Institute of Technology Madras}\\
{\tt\small ai23resch04001@iith.ac.in \quad aravind@cerai.in \quad krmopuri@ai.iith.ac.in}
}

\begin{document}

\wacvacceptedwatermark
\maketitle

\begin{abstract}
In \textit{Dataset Condensation}, the goal is to synthesize a small dataset that replicates the training utility of a large original dataset. Existing condensation methods synthesize datasets with significant redundancy, so there is a dire need to reduce redundancy and improve the diversity of the synthesized datasets. To tackle this, we propose an intuitive \textbf{Di}versity \textbf{Re}gularizer (\textbf{DiRe}) composed of cosine similarity and Euclidean distance, which can be applied off-the-shelf to various state-of-the-art condensation methods. Through extensive experiments, we demonstrate that the addition of our regularizer improves state-of-the-art condensation methods on various benchmark datasets from CIFAR-10 to ImageNet-1K with respect to generalization and diversity metrics.
\end{abstract}
\section{Introduction}\label{sec:introduction}
Training datasets for modern neural networks have grown to large scales, making the learning process cumbersome and expensive. To ameliorate this issue, there has been tremendous recent interest in \textit{Dataset Condensation} (DC), also referred to as \textit{Dataset Distillation}~\citep{wang2018dataset,zhao2021dataset,sachdeva2023data,yang2024what,cui2024ameliorate,qi2024fetch,ding2024condtsf,qin2024a,shao2024elucidating}. Here, the goal is to generate a small synthetic dataset from the original large dataset that can instead be used for neural network training and other related tasks, such as neural architecture search~\citep{such2020generative}. Furthermore, dataset condensation has been shown to have several other applications, such as memory rehearsal in continual learning~\citep{deng2022remember,gu2024summarizing}, data privacy~\citep{dong2022privacy,chung2024rethinking,loo2024understanding}, and federated learning~\citep{xiong2023feddm,huang2024overcoming}.

Other active approaches for data-efficient deep learning include Data Subset Selection~\citep{kai2015submodularity,jain2023efficient}, Coreset Selection~\citep{har2004coresets,yang2023towards,xia2023moderate}, and Dataset Pruning~\citep{yang2023dataset,zhang2023selectivity,zhang2024spanning}. In all these approaches, a \textit{subset} of the original dataset is selected as a substitute for model training. Since they are constrained to only select \textit{real} data points from the training dataset, they typically need to produce much larger datasets than DC approaches for comparable performance. 

Most work on DC has focused on bi-level optimization-based strategies, where the outer optimization task is for synthetic dataset updates, and the inner optimization task is for model updates. Following the taxonomy in ~\citep{sachdeva2023data}, we note that there are primarily four different high-level approaches to DC, such as those focusing on meta-model matching~\citep{wang2018dataset,zhou2022dataset}, gradient matching~\citep{zhao2021dataset, zhao2021siamese}, trajectory matching~\citep{cazenavette2022cvpr,guo2024towards,lee2024selmatch}, and distribution matching~\citep{zhao2023distribution,zhao2023improved}.

\begin{figure}[t!]
  \centering
  \includegraphics[width=\linewidth]{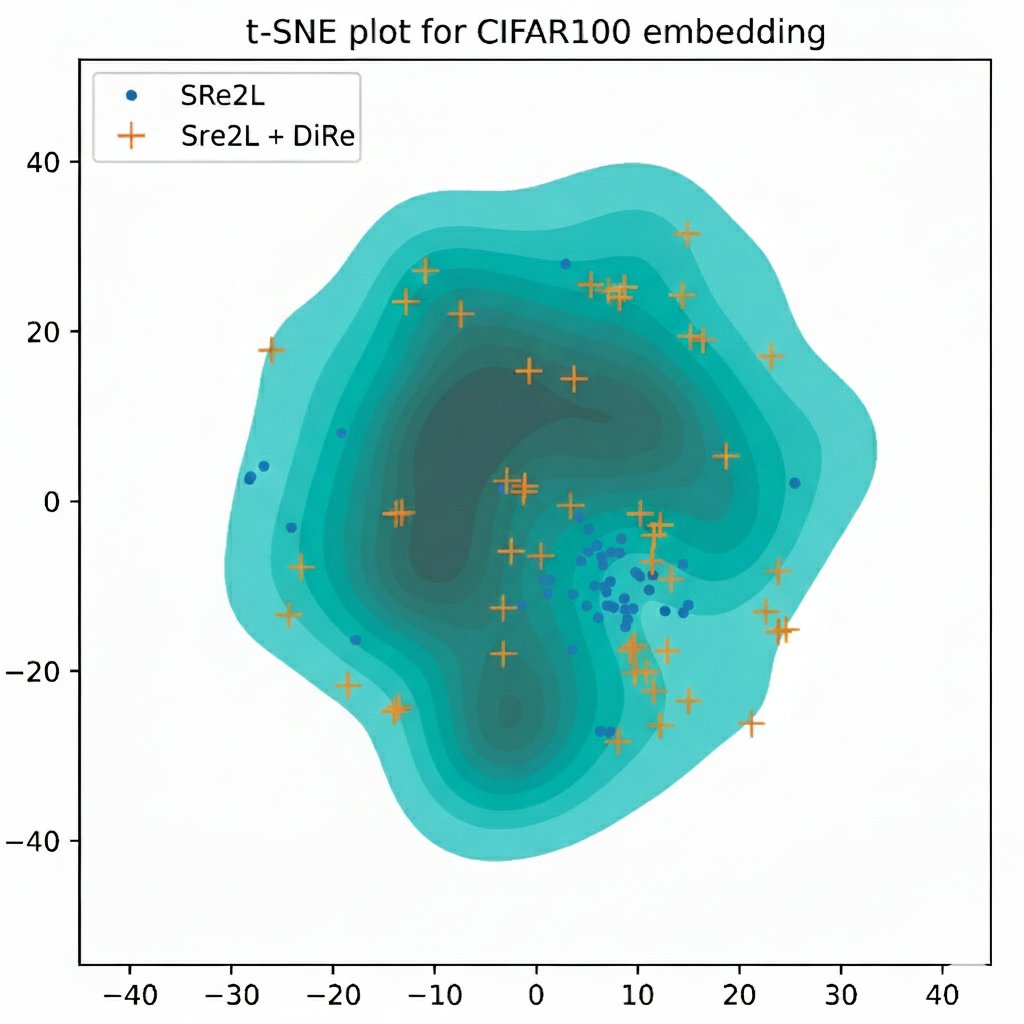}
  \caption{t-SNE plot of embeddings of synthetic images generated from the CIFAR100 dataset with IPC=50 settings. As can be seen, synthetic images generated by adding the \textbf{DiRe} regularizer are more diverse than the vanilla SRe$^2$L~\citep{yinsqueeze} method. The synthetic images are spread throughout the original dataset's feature space (shown in Cyan color).}
  \label{fig:tsne_plot}
\end{figure}
Before 2023, due to the bi-level optimization nature of most DC algorithms, none of them could scale to large-scale datasets such as ImageNet-1K~\cite{russakovsky2015imagenet}, as they require storing the entire original training dataset in memory. To circumvent this issue,~\citet{yinsqueeze} proposed SRe$^2$L, which decouples the outer and inner optimization tasks, enabling DC on large datasets. Recent works have been based on such decoupled optimization-based strategies~\citep{yin2024cda,shao2024elucidating,du2024diversity,zhang2025breaking,delt_2025_CVPR,shao2024generalized}.

Since the primary goal of DC is to reduce redundancy, it is natural to ask whether explicitly encouraging diversity in the synthesized dataset is beneficial. Surprisingly, very little prior work has focused on diversity in dataset condensation~\citep{du2024diversity,sun2024diversity}, despite \textit{dataset diversity} having been widely acknowledged as a crucial aspect for the successful training of machine learning models~\citep{diversityml2019,van2021benchmarking,yang2021just,zhao2024position}. The diversity-driven DC approaches~\citep{du2024diversity,sun2024diversity} currently offer state-of-the-art performance for dataset condensation. This motivates us to ask the following natural question: 

{\em Can we use a simple and intuitive regularizer to enhance diversity in dataset condensation, leading to better generalization?}

Surprisingly, we found that the answer is a resounding ``Yes!''. We initially started using only \textit{cosine similarity} as a regularizer, but found that augmenting it with \textit{Euclidean distance} helps significantly. Thus, combining these terms led us to our \textbf{Di}versity \textbf{Re}gularizer \textbf{DiRe}, which can be applied off-the-shelf to any existing optimization-based DC algorithm to improve the diversity (and thus generalization) of distilled datasets.

Another key issue we noticed is that both \cite{du2024diversity} and \cite{sun2024diversity} face a pitfall, which is very common to ML research studying diversity. This pitfall is highlighted vividly by the title of ~\citep{zhao2024position}, which won one of the outstanding paper awards in ICML 2024; ``Measure Dataset Diversity, Don't Just Claim It''. Although \cite{du2024diversity} and \cite{sun2024diversity} \textit{claim} that they produce datasets that are more diverse than prior DC algorithms, they do not quantitatively measure diversity using any well-established metrics. We also tackle this by studying the diversity of distilled datasets using established quantitative notions of diversity~\citep{naeem2020reliable,friedman2023the}. 
Our main contributions are as follows.
\begin{itemize}
    \item We propose an intuitive \textbf{Di}versity \textbf{Re}gularizer (\textbf{DiRe}) that can be applied off-the-shelf to any dataset condensation algorithm that has a separate synthesis stage.
    \item We are the first to quantitatively study the diversity of distilled datasets using well-established Dataset Diversity measures such as Coverage~\citep{naeem2020reliable} and Vendi Score~\citep{friedman2023the}. We show that adding DiRe significantly improves SRe$^2$L on both these metrics.
    \item We also demonstrate that adding DiRe to several other optimization-based DC methods improves their performance across generalization and diversity measures. More specifically, we consider DWA~\citep{du2024diversity}, CDA~\citep{yin2024cda}, UFC~\citep{zhang2025breaking}, DELT~\citep{delt_2025_CVPR}, G-VBSM~\citep{shao2024generalized}, MTT~\citep{cazenavette2022cvpr}, and DM~\citep{zhao2023distribution}. Note that MTT and DM are trajectory matching and distribution matching DC algorithms, which are very different from decoupling-based algorithms such as SRe$^2$L, DWA, CDA, UFC, DELT, and G-VBSM.
\end{itemize}

We would like to highlight that our proposed regularizer is broadly applicable to all optimization-based DC methods, which constitute the majority of all DC algorithms. But, there are a few DC algorithms such as RDED~\citep{sun2024diversity}, to which DiRe cannot be added as-is. These algorithms do not incorporate a separate synthesis stage where optimization occurs.
\section{Methodology}
\label{sec:methodology}
In this section, we describe our diversity regularizer DiRe in detail and provide Algorithm~\ref{alg:DiRe} which shows how DiRe can be added to SRe$^2$L. Note that DiRe can be added to other optimization-based DC methods in a very similar fashion. Table~\ref{table:notations} provides the notation used in the rest of the paper.

\begin{table}[hbt!]
\caption{Notation}
\label{table:notations}
\begin{tabularx}{\columnwidth}{lX}
    \toprule
    Symbol  & Description \\
    \midrule
    $V$     & Original large-scale training dataset \\
    $S$     & Distilled synthetic dataset \\
    $\theta$ & Parameters of the deep neural network \\
    $f_\theta$ & Pretrained teacher network \\
    $h_\theta$ & Feature extractor of $f_\theta$ \\
    $\eta$ & Learning rate for distillation \\
    $r_c$ & Weight for pairwise cosine similarity loss \\
    $r_e$ & Weight for pairwise Euclidean distance loss \\
    $S_{cos}(A,B)$ & Pairwise cosine similarity between rows of matrices $A$ and $B$ \\
    $D_{euc}(A,B)$ & Pairwise Euclidean distance between rows of matrices $A$ and $B$ \\
    $L_{ce}$ & Cross-entropy loss \\
    $L_{bn}$ & Batch-Norm loss \\
    $\langle x,y \rangle$ & Inner product of vector $x$ and vector $y$ \\
    $\|x\|$ & $\ell_2$-norm (Euclidean-norm) of vector $x$ \\
    \bottomrule
\end{tabularx}
\end{table}

Given a large original labeled dataset $V =\{(x_i,y_i)\}_{i=1}^{|V|}$, the aim of dataset condensation is to condense it to a much smaller synthetic dataset $S = \{\tilde{x}_i,\tilde{y}_i\}_{i=1}^{|S|}$, such that a model $\theta^S$ trained on $S$ has a generalization performance close to that of a model $\theta^V$ trained on $V$.

Most DC algorithms involve solving a bi-level optimization problem with the outer loop for updating the synthetic dataset and the inner loop for updating the model.~\citet{yinsqueeze} decoupled this bi-level optimization and introduced SRe$^2$L, a lightweight tripartite learning paradigm that can scale to large datasets such as ImageNet-1K. The three stages of SRe$^2$L~\citep{yinsqueeze} are:
\begin{itemize}
\item \textbf{S}queeze, in which a model is trained on the original training dataset, and the Batch-Norm (BN) layer statistics are stored.\item \textbf{Re}cover, in which the BN statistics are used to synthesize the condensed dataset.
\item \textbf{Re}label, in which the synthetic dataset is assigned soft labels through the model trained on the original dataset.
\end{itemize}

The synthesis stage (Recover) of SRe$^2$L consists of synthesizing the target data starting from independent random Gaussian noise.~\citet{yinsqueeze} implicitly assumes that starting from random Gaussian noise provides diversity in this parallelized synthesis process. However, as shown in Figure \ref{fig:tsne_plot}, synthetic datasets generated by SRe$^2$L cover only a small portion of the entire dataset manifold.~\citet{du2024diversity} improve the diversity of SRe$^2$L by carefully tailoring their synthesis process using the specific properties of the Batch-Normalization statistics. In contrast to their approach, we show that a simple and intuitive diversity regularizer based on cosine similarity and Euclidean distances is sufficient to ensure diversity, leading to better generalization.

DiRe consists of three components. 
\begin{enumerate}
    \item \textbf{Cosine Diversity loss (CD)}: Promotes diversity among the synthetic dataset by minimizing the pairwise cosine similarities between their embeddings, encouraging a more dispersed representation in the embedding space.
    \item \textbf{Cosine Distribution Matching loss (CDM)}: Encourages the synthetic data directions to align with the real data in the embedding space by maximizing the pairwise cosine similarity between synthetic and real image embeddings.
    \item \textbf{Euclidean Distribution Matching loss (EDM)}: Encourages synthetic embeddings to cluster near real ones in Euclidean space by minimizing the pairwise Euclidean distances between synthetic and real embeddings.
\end{enumerate}

Let $X_{syn}^{c}$ be the set of all the synthetic images belonging to class $c$ and $X_{real}^{c}$ be the set of all the real images belonging to class $c$. Embeddings of the synthetic images computed through the pre-trained feature extractor network are given as $E_{syn}^{c} = h_\theta(X_{syn}^{c})$ and embeddings of real images are given as $E_{real}^{c} = h_\theta(X_{real}^{c})$.

Pairwise cosine similarity between two matrices $A$ and $B$ with dimensions $N \times D$ and $M \times D$ (i.e., set of N and M vectors of D dimensions respectively; $A^i$ is the $i^{th}$ row in $A$) 
 is calculated as shown in Equation \ref{eq:pairwise_cosine_eq}.
\begin{equation}
\displaystyle S_{cos}(A,B) =\sum_{i=1}^{N}{\sum_{j=1}^{M}{\frac{\sum_{d=1}^{D}A_d^i \cdot B_d^j}{\sqrt{\sum_{d=1}^{D}(A_d^i)^2}\cdot \sqrt{\sum_{d=1}^{D}(B_d^j)^2}}}}
\label{eq:pairwise_cosine_eq}
\end{equation}

Pairwise Euclidean distance between two matrices $A$ and $B$ with $D$ number of features is calculated as shown in Equation \ref{eq:pairwise_euclidean_distance}. 
\begin{equation}
\displaystyle D_{euc}(A,B) = \sum_{i=1}^{N}{\sum_{j=1}^{M}\sqrt{\sum_{d=1}^{D}(A_d^i-B_d^j)^2}}
\label{eq:pairwise_euclidean_distance}
\end{equation}

The three components of DiRe are formulated as follows.
\begin{equation}
\displaystyle \text{CD} = l_{cos\_div}^c= S_{cos}(E_{syn}^c,E_{syn}^c)
\label{eq:first_regularizer}
\end{equation}

\begin{equation}
\displaystyle \text{CDM} = l_{cos\_dm}^c= 1-S_{cos}(E_{syn}^c,E_{real}^c)
\label{eq:second_regularizer}
\end{equation}

\begin{equation}
\text{EDM} = l_{euc\_dm}^c = D_{euc}(E_{syn}^c,E_{real}^c)
\label{eq:third_regularizer}
\end{equation}

Note that CD encourages synthetic examples to spread out, CDM pushes them toward real data directions, and EDM brings them close to real data in Euclidean distance. Algorithm \ref{alg:DiRe} presents our approach more formally.

\begin{algorithm}[htb!]
\caption{SRe$^2$L with Diversity Regularizer (DiRe)}
\label{alg:DiRe}
\begin{algorithmic}[1]
\Require Images per class $ipc$, feature extractor network $h_\theta$, number of iterations $T$, learning rate $\eta$, cosine distance weight $r_c$, Euclidean distance weight $r_e$, Number of classes $C$
\Ensure Distilled dataset $S_T$
\State Initialize $S_0$ with $C \times ipc$ images drawn from a Gaussian noise distribution
\State Forward pass real data through $h_\theta$, Store real embeddings $E_{real}^c$ $\forall c \in \{0,\ldots C-1\}$

\For{$t=1$ to $T$}
\State Forward pass synthetic data $S_{t-1}$ through $h_\theta$
\State Obtain syn. embeddings $E_{syn}^c$ $\forall c \in \{0,\ldots C-1\}$
\State Compute $l_{\text{cos\_div}}^c$, $l_{\text{cos\_dm}}^c$, and $l_{\text{euc\_dm}}^c$
\State $L_{\text{syn}} = \sum_{c=1}^C r_c \cdot (l_{\text{cos\_div}}^c + l_{\text{cos\_dm}}^c) + r_e\cdot l_{\text{euc\_dm}}^c$
\State $L_{\text{total}} \gets L_{\text{ce}} + L_{\text{bn}} + L_{\text{syn}}$
\State $S_t \gets S_{t-1} - \eta \cdot \nabla_S L_{\text{total}}$
\EndFor
\end{algorithmic}
\end{algorithm}
\section{Implementation of diversity regularizer}
\subsection{Embeddings w.r.t. feature extraction layer}
We use the outputs of the penultimate layers (for example, the output of the Average Pool layer of ResNet-18) as the feature-rich, low-dimensional representations of the real and synthetic images. These embeddings are used to compute the components of DiRe. For ResNet-18, this yields 512-dimensional features. 

\subsection{Pairwise cosine similarity and pairwise Euclidean distance}
The complexity of computing pairwise cosine similarity (Equation \ref{eq:pairwise_cosine_eq}) and pairwise Euclidean distance (Equation \ref{eq:pairwise_euclidean_distance}) between two matrices of $K$ rows (i.e., computation between embeddings of $K$ images) is $\mathcal{O}(K^2)$. Carrying out these computations in a nested loop manner for every epoch and every class would be prohibitively expensive. Instead, we utilize these functions from the torchmetrics library\footnote{\url{https://lightning.ai/docs/torchmetrics/stable/pairwise/cosine_similarity.html}}. Because of their efficient implementation, they achieve $\approx 30$x faster computation. The timing comparison is provided in the subsection \ref{timing_analysis}.
\section{Experiments and results}
\label{sec:Experiments}
\subsection{Experimental Setup}
\textbf{Applications}. We evaluate the performance of our method on image classification. 

\textbf{Datasets}. For image classification, we evaluate the effectiveness of our regularizer on four popular benchmark image classification datasets, i.e., CIFAR-10, CIFAR-100~\citep{krizhevsky2009learning}, Tiny ImageNet~\citep{Le2015TinyIV}, and ImageNet-1K~\citep{russakovsky2015imagenet}. To test the robustness of our method, we consider all these datasets with the number of classes varying from $10$ to $1000$.

\textbf{Backbone architecture}. Similar to other SOTA works, we use the ResNet-18~\citep{he2016deep} architecture to condense the datasets. Unless specified otherwise, we use the same trained model as a teacher model for carrying out knowledge distillation. We also use other CNN architectures, such as ResNet-50~\citep{he2016deep}, ResNet-101~\citep{he2016deep}, VGG-16~\citep{Simonyan2015VeryDC}, and MobileNetV2~\citep{MobileNetV2_2018_CVPR}, in addition to the transformer architecture ViT~\citep{lee2021vision}, to carry out our cross-architecture generalization study. The main aim is to study the effect of improved diversity on generalization over the backbone architecture and cross-architecture generalization over various architectures. 

\textbf{Diversity Metrics}. We consider Coverage~\citep{naeem2020reliable}, Vendi Score~\citep{friedman2023the}, and intra-class cosine similarity as the diversity metrics. \textbf{Coverage} measures the fraction of real samples whose neighbourhoods contain at least one synthetic sample. It is calculated as: 
\begin{equation}
\text{coverage} = \frac{1}{N} \sum_{i=1}^{N} 1_{\left\{ \exists\, j \ \text{s.t.} \ Y_j \in B\big(X_i, \text{NND}_k(X_i)\big) \right\}}
\label{eq:coverage}
\end{equation}

where, $X$ is the real dataset, $Y$ is the synthetic dataset, NND$_k$ represents $k^{th}$ Nearest Neighbor Distance, and $B(X,r)$ represents a ball of radius $r$ around the data point $X$. All the calculations are carried out in the feature space, using embeddings of the \textit{avgpool} layer of the ResNet-18 architecture. Hence, a higher coverage value indicates higher diversity among the synthetic images. 

\textbf{Vendi Score} is defined as the exponential of the Shannon
entropy of the eigenvalues of a similarity matrix. A higher Vendi Score indicates greater diversity in the dataset. 

\textbf{Cosine similarity} between two vectors $x$ and $y$ is:
\begin{equation}
\displaystyle s_{cos}(x,y) = \frac{ \langle x,y \rangle }{ \|x\| \cdot \|y\|}
\label{eq:cosine_sim}
\end{equation}
We compute intra-class pairwise cosine similarity among the embeddings of the synthetic dataset and use the mean as a measure of diversity. Lower intra-class cosine similarity indicates higher diversity. 

\textbf{Baselines} We consider the results reported by SRe$^2$L, CDA, DWA, UFC, G-VBSM, and DELT wherever available. We reuse the publicly available codebases for each method to generate results where they are unavailable. The same hyperparameters are used for all methods to ensure uniformity. 

To further demonstrate the generality of our approach, we apply it to two additional optimization-based dataset condensation methods: Matching Training Trajectory~\citep{cazenavette2022cvpr} and Distribution Matching~\citep{zhao2021dataset}. We use the original MTT codebase and the DC-BENCH~\footnote{\url{https://github.com/justincui03/dc_benchmark}} implementation for DM.

\textbf{Codebase} We make our codebase publicly available for easy reproduction of our results\footnote{\url{https://github.com/DIL-IITH/DiRe}}.

\subsection{Results for CIFAR-10}
A comparison of the accuracy values and diversity metrics obtained on a randomly initialized ResNet-18 architecture, trained through knowledge distillation from a pre-trained ResNet-18 model, is presented in Tables~\ref{cifar10_results_acc} and~\ref{cifar10_results_div}. We see that the addition of DiRe leads to improvements in generalization accuracy and diversity metrics. 

\begin{table}[hbt!]
\caption{Impact on accuracy by addition of DiRe to various DC methods on CIFAR-10. It can be seen that the addition of DiRe leads to an increase in the accuracy obtained by each of the methods considered.}
\label{cifar10_results_acc}
\centering
\resizebox{\columnwidth}{!}{%
\begin{tabular}{lccc}
\toprule
Methods & IPC$=10$ & IPC$=50$ & IPC$=100$ \\
\toprule 
SRe$^2$L& 27.2 {\textpm} 0.4 & 47.5 {\textpm} 0.5 & 57.5 {\textpm} 0.6 \\
SRe$^2$L + DiRe&\bf 37.4 {\textpm} 1.1&\bf 59.7 {\textpm} 1.2 &\bf 71.2 {\textpm} 1.2 \\
\midrule
DWA& 32.6 {\textpm} 0.4 & 53.1 {\textpm} 0.3 & 67.2 {\textpm} 0.3 \\
DWA + DiRe&\bf 36.5 {\textpm} 0.9 &\bf 62.2 {\textpm} 0.7 & \bf 71.0 {\textpm} 0.6 \\
\midrule 
INFER&32.0 {\textpm} 0.5 & 60.4 {\textpm} 1.6 & - \\
INFER + DiRe&\bf 45.3 {\textpm} 0.8&\bf 73.9 {\textpm} 0.2& - \\
\midrule 
INFER (D)& 30.7 {\textpm} 0.3&60.7 {\textpm} 0.9& - \\
INFER(D) + DiRe&\bf 57.1 {\textpm} 0.9 &\bf 85.1 {\textpm} 0.3 & - \\
\midrule 
G-VBSM & 53.5 {\textpm} 0.6 & 59.2 {\textpm} 0.4 & - \\
G-VBSM + DiRe & \bf 55.8 {\textpm} 0.2 & \bf 68.3 {\textpm} 0.3 & - \\
\midrule 
DELT & 43.0 {\textpm} 0.9 & 64.9 {\textpm} 0.9 & - \\
DELT + DiRe & \bf 49.2 {\textpm} 0.5 & \bf 76.3 {\textpm} 0.2 & -\\

\bottomrule
\end{tabular}}
\end{table}

\begin{table}[hbt!]
\caption{Impact on diversity by the addition of DiRe to various DC methods on CIFAR-10. The addition of DiRe has resulted in the generation of a synthetic dataset with higher diversity.}
\label{cifar10_results_div}
\centering
\resizebox{\columnwidth}{!}{%
\begin{tabular}{lccc}
\toprule
Methods & Coverage $\uparrow$ & Similarity $\downarrow$ & Vendi $\uparrow$ \\
\toprule 
SRe$^2$L& 2.25\% & 0.90 & 1.87 \\
SRe$^2$L + DiRe&\bf 3.53\% & \bf 0.86 & \bf 2.25 \\
\midrule
DWA& 2.43\% & 0.88 & 2.20 \\
DWA + DiRe&\bf 2.84\% & \bf 0.78 & \bf 2.34 \\
\midrule 
INFER& 2.97\% & \bf 0.74 & 2.02 \\
INFER + DiRe& \bf 6.74\% & 0.82 & \bf 2.24 \\
\midrule 
INFER (D)& 2.97\% & \bf 0.74 & 2.02 \\
INFER(D) + DiRe& \bf 6.74\% & 0.82 & \bf 2.24 \\
\midrule 
G-VBSM & 0.04$\%$ & 0.76 & 2.07 \\
G-VBSM + DiRe & \bf 0.06$\%$ & \bf 0.69 & \bf 2.67 \\
\midrule 
DELT & 0.9$\%$ & \bf 0.84 & 1.69 \\
DELT + DiRe & \bf 4.3$\%$ & 0.93 & \bf 2.28 \\

\bottomrule 
\end{tabular}}
\end{table}

\subsection{Results for CIFAR-100}
Tables~\ref{cifar100_results_acc} and~\ref{cifar100_results_div} compare accuracy values and diversity metrics for various methods at various IPC settings. DiRe is able to improve both accuracy and diversity metrics for all the DC methods considered.

\begin{table}[hbt!]
\caption{Impact on accuracy by addition of DiRe to various DC methods on CIFAR-100. It can be seen that the addition of DiRe leads to an increase in the accuracy obtained by each of the methods considered.}
\label{cifar100_results_acc}
\centering
\resizebox{\columnwidth}{!}{%
\begin{tabular}{lccc}
\toprule
Methods & IPC$=10$ & IPC$=50$ & IPC$=100$ \\
\toprule 
SRe$^2$L& 31.6 {\textpm} 0.5 & 52.2 {\textpm} 0.3 & 57.5 {\textpm} 0.6 \\
SRe$^2$L + DiRe& \bf 41.2 {\textpm} 1.1&\bf 63.4 {\textpm} 0.2 &\bf 66.5 {\textpm} 0.2 \\
\midrule
DWA& 39.6 {\textpm} 0.6 & 60.9 {\textpm} 0.5 & 65.2 {\textpm} 0.3 \\
DWA + DiRe&\bf 41.4 {\textpm} 0.4 &\bf 62.3 {\textpm} 0.2 & \bf 65.3 {\textpm} 0.2 \\
\midrule 
CDA & 49.8 {\textpm} 0.6 & 64.4 {\textpm} 0.5 & 65.5 {\textpm} 0.1 \\
CDA + DiRe & \bf 54.5 {\textpm} 0.3 &\bf 66.6 {\textpm} 0.1 & \bf 68.0 {\textpm} 0.4 \\
\midrule
INFER&45.2 {\textpm} 0.1 & 62.8 {\textpm} 0.4 & 66.3 {\textpm} 0.1 \\
INFER + DiRe&\bf 53.7 {\textpm} 1.5&\bf 67.6 {\textpm} 0.2& \bf 69.2 {\textpm} 0.3 \\
\midrule 
INFER (D)& 53.4 {\textpm} 0.6&68.9 {\textpm} 0.1& 73.3 {\textpm} 0.2 \\
INFER(D) + DiRe&\bf 63.9 {\textpm} 0.2 &\bf 74.1 {\textpm} 0.1 & \bf 76.1 {\textpm} 0.2 \\
\bottomrule
\end{tabular}}
\end{table}

\begin{table}[hbt!]
\caption{Impact on diversity by the addition of DiRe to various DC methods on CIFAR-100. Addition of DiRe results in improved diversity across all the DC methods considered.}
\label{cifar100_results_div}
\centering
\resizebox{\columnwidth}{!}{%
\begin{tabular}{lccc}
\toprule
Methods & Coverage $\uparrow$ & Similarity $\downarrow$ & Vendi $\uparrow$ \\
\toprule 
SRe$^2$L& 14.87\% & 0.81 & 2.79 \\
SRe$^2$L + DiRe&\bf 23.12\% & \bf 0.65 & \bf 3.08\\
\midrule
DWA& 19.32\% & 0.77 & 2.99 \\
DWA + DiRe&\bf 32.75\% & \bf 0.61 & \bf 3.11 \\
\midrule 
CDA & 12.31\% & 0.81 & 2.43 \\
CDA + DiRe & \bf 15.43\% & \bf 0.67 & \bf 2.85 \\
\midrule
INFER& 14.30\% & 0.68 & 2.66 \\
INFER + DiRe& \bf 28.99\% & \bf 0.60 & \bf 2.70 \\
\midrule 
INFER (D)& 14.30\% & 0.68 & 2.66 \\
INFER(D) + DiRe& \bf 28.99\% & \bf 0.60 & \bf 2.70 \\
\bottomrule 
\end{tabular}}
\end{table}

\begin{figure}[hbt!]
\begin{center}
\centerline{\includegraphics[width=\columnwidth]{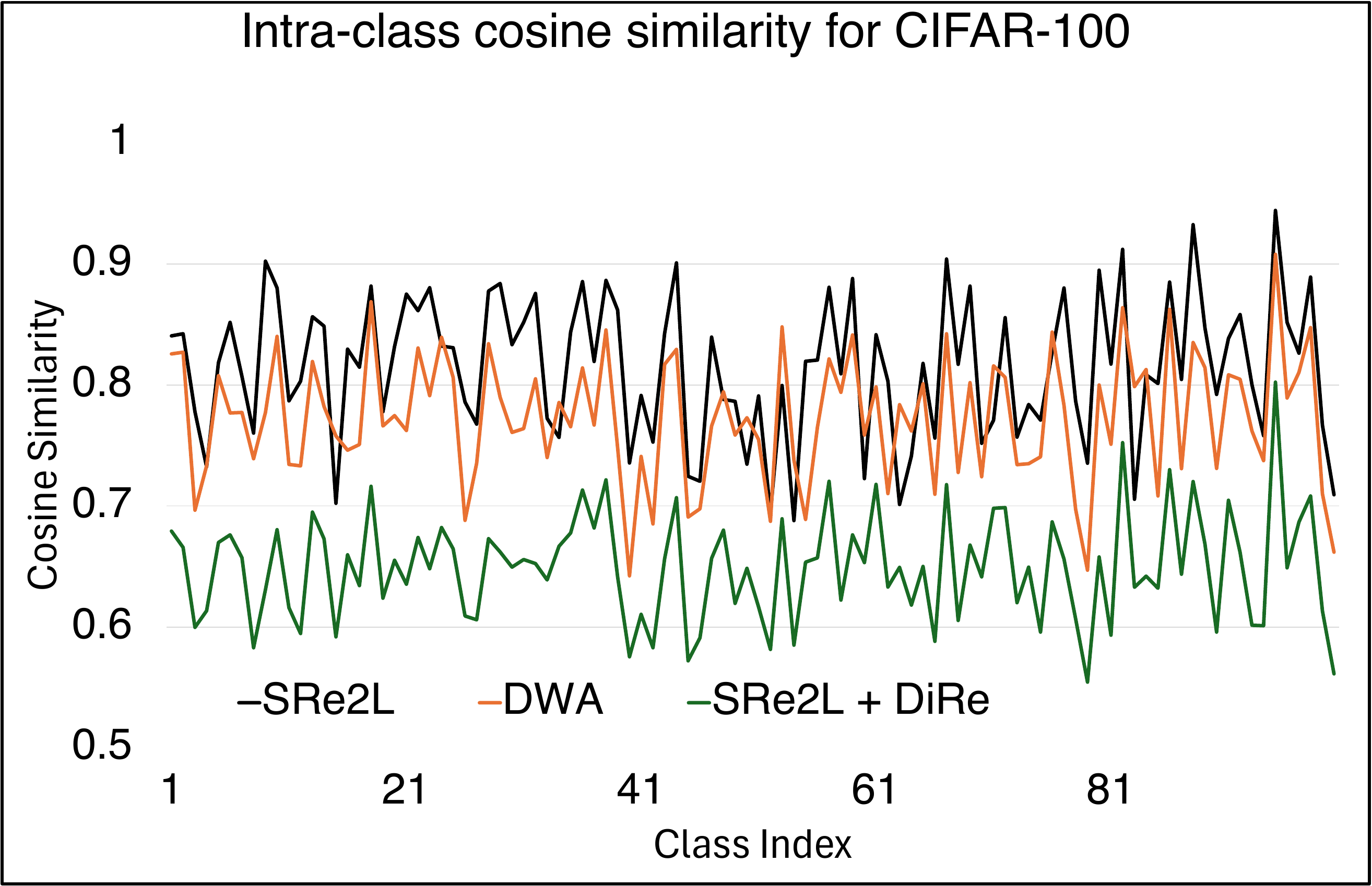}}
\caption{Class-wise intra-class cosine similarity for CIFAR-100. Lower cosine similarity indicates higher diversity among the synthetic dataset. SRe$^2$L + DiRe clearly shows lower cosine similarity for all the classes in the dataset as compared to vanilla SRe$^2$L and DWA.}
\label{fig:cosine_similarity_fig}
\end{center}
\end{figure}

The plot of class-wise intra-class cosine similarity is shown in Figure \ref{fig:cosine_similarity_fig}, which further showcases the diversity introduced by our method. t-SNE plot for embeddings of CIFAR-100 for IPC=50 setting is shown in Figure \ref{fig:tsne_plot}. As we can see, our method can cover more diverse regions in the original data manifold compared to SRe$^2$L and DWA.

\subsection{Results for Tiny ImageNet}
Tables~\ref{tiny_results_acc} and~\ref{tiny_results_div} compare the accuracy values and diversity metrics for various methods on Tiny ImageNet trained on ResNet-18. DiRe improves both generalization accuracy and diversity metrics for all three condensation methods considered.  

\begin{table}[hbt!]
\caption{Impact on accuracy by addition of DiRe to various DC methods on Tiny ImageNet. It can be seen that the addition of DiRe leads to an increase in the accuracy values obtained by each of the methods considered.}
\label{tiny_results_acc}
\centering
\resizebox{\columnwidth}{!}{%
\begin{tabular}{lccc}
\toprule
Methods & IPC$=10$ & IPC$=50$ & IPC$=100$ \\
\toprule 
SRe$^2$L& 17.7 {\textpm} 0.7 & 41.1 {\textpm} 0.4 & 49.7 {\textpm} 0.3 \\
SRe$^2$L + DiRe& \bf 34.7 {\textpm} 0.3&\bf 55.3 {\textpm} 0.4 &\bf 57.4 {\textpm} 0.1 \\
\midrule
DWA& 32.1 {\textpm} 0.1 & 52.8 {\textpm} 0.2 & 56.0 {\textpm} 0.2 \\
DWA + DiRe& \bf 37.6 {\textpm} 0.3 & \bf 55.2 {\textpm} 0.1 & \bf 58.5 {\textpm} 0.2 \\
\midrule 
CDA & 21.3 {\textpm} 0.3 & 48.7 {\textpm} 0.1 & 53.2 {\textpm} 0.1 \\
CDA + DiRe &  \bf 34.8 {\textpm} 0.5& \bf 54.5 {\textpm} 0.2 & \bf 56.5 {\textpm} 0.3 \\
\midrule 
G-VBSM & - & 47.6 {\textpm} 0.3 & 51.0 {\textpm} 0.4 \\
G-VBSM + DiRe & - & \bf 50.1 {\textpm} 0.2 & \bf 54.2 {\textpm} 0.5 \\
\midrule 
DELT & 43.0 {\textpm} 0.1 & 55.7 {\textpm} 0.5 & - \\
DELT + DiRe & \bf 45.7 {\textpm} 0.5 & \bf 56.8 {\textpm} 0.1 & - \\
\bottomrule
\end{tabular}}
\end{table}

\begin{table}[hbt!]
\caption{Impact on diversity by the addition of DiRe to various DC methods on Tiny ImageNet. Diversity metrics across the three different metrics improve after the addition of DiRe.}
\label{tiny_results_div}
\centering
\resizebox{\columnwidth}{!}{%
\begin{tabular}{lccc}
\toprule
Methods & Coverage $\uparrow$ & Similarity $\downarrow$ & Vendi $\uparrow$\\
\toprule 
SRe$^2$L& 30\% & \bf 0.66 & 3.07 \\
SRe$^2$L + DiRe&\bf 45\% & \bf 0.66 & \bf 3.22\\
\midrule
DWA& 36\% & 0.69 & 3.04 \\
DWA + DiRe&\bf 52\% & \bf 0.65 & \bf 3.14 \\
\midrule 
CDA & 32\% & 0.75 & 6.41 \\
CDA + DiRe & \bf 53\% & \bf 0.69 & \bf 6.96 \\
\midrule 
G-VBSM & 36$\%$ & 0.71 & 2.63 \\
G-VBSM + DiRe & \bf 45.5$\%$ & \bf 0.66 & \bf 3.61 \\
\midrule 
DELT & 6.5 $\%$ & 0.62 & 2.18 \\
DELT + DiRe & \bf 8.2$\%$ & \bf 0.59 & \bf 2.43 \\
\bottomrule
\end{tabular}}
\end{table}

\subsection{Results for ImageNet-1K}
\begin{figure*}[hbt!]
\begin{center}
\includegraphics[width=\linewidth]{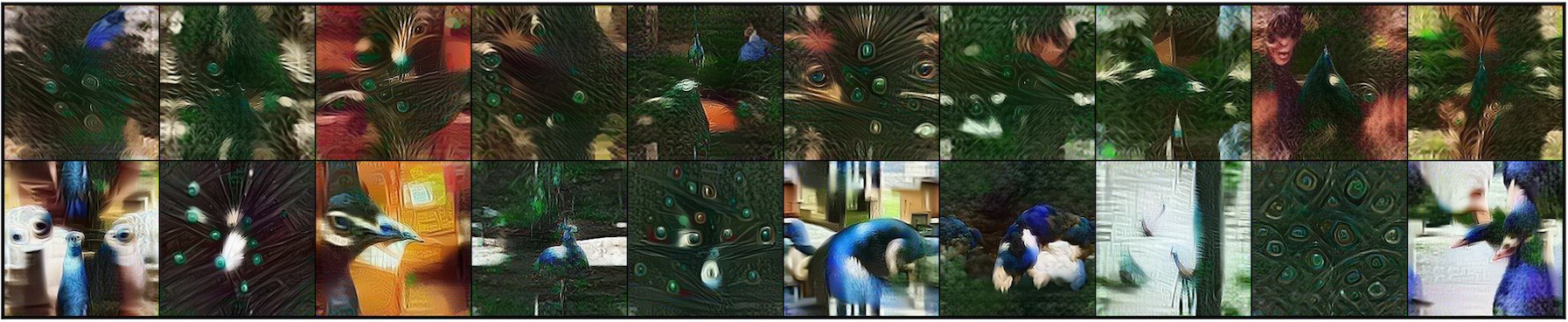}
\caption{Visualization of synthetic images belonging to the `Peacock' class from the ImageNet-1K dataset. The top row images are condensed through SRe$^2$L, and the bottom row images are condensed through SRe$^2$L + DiRe. The diversity among the synthetic images is clearly visible.}
\label{ilsvrc_visualization}
\end{center}
\end{figure*}

A comparison of accuracy values and diversity metrics on a synthetic dataset generated from ImageNet-1K is presented in Tables~\ref{ilsvrc_results_acc} and~\ref{ilsvrc_results_div}. The addition of DiRe results in an improvement in both accuracy and diversity metrics. Figure~\ref{ilsvrc_visualization} showcases synthetic images belonging to the `Peacock' class condensed from the ImageNet-1K dataset. Diversity among the synthetic images resulting from the addition of DiRe is clearly noticeable.

\begin{table}[hbt!]
\caption{Impact on accuracy by addition of DiRe to various DC methods on ImageNet-1K. It can be seen that the addition of DiRe leads to an increase in the accuracy obtained by each of the methods considered.}
\label{ilsvrc_results_acc}
\centering
\resizebox{\columnwidth}{!}{%
\begin{tabular}{lccc}
\toprule
Methods & IPC$=10$ & IPC$=50$ & IPC$=100$ \\
\toprule 
SRe$^2$L& 21.3 {\textpm} 0.6 & 46.8 {\textpm} 0.2 & 52.8 {\textpm} 0.4 \\
SRe$^2$L + DiRe& \bf 38.5 {\textpm} 0.1&\bf 55.6 {\textpm} 0.3 &\bf 59.2 {\textpm} 0.1 \\
\midrule
DWA& 37.9 {\textpm} 0.2 & 55.2 {\textpm} 0.2 & 59.2 {\textpm} 0.3 \\
DWA + DiRe& \bf 39.1 {\textpm} 0.4 & \bf 56.9 {\textpm} 0.1 & \bf 61.0 {\textpm} 0.1 \\
\midrule 
CDA & 33.5 {\textpm} 0.3 & 52.5 {\textpm} 0.3 & 58.0 {\textpm} 0.2 \\
CDA + DiRe & \bf 35.6 {\textpm} 0.1& \bf 56.0 {\textpm} 0.1 & \bf 60.3 {\textpm} 0.2 \\
\midrule
INFER&28.7 {\textpm} 0.2 & 51.8 {\textpm} 0.2 & - \\
INFER + DiRe&\bf 38.2 {\textpm} 0.3&\bf 61.2 {\textpm} 0.5& - \\
\midrule 
G-VBSM & 31.4 {\textpm} 0.5 & 51.8 {\textpm} 0.4 & 55.7 {\textpm} 0.4 \\
G-VBSM + DiRe & \bf 35.1 {\textpm} 0.1 & \bf 55.2 {\textpm} 0.2 & \bf 58.7 {\textpm} 0.1 \\
\midrule 
DELT & 46.1 {\textpm} 0.4 & 59.2 {\textpm} 0.4 & - \\
DELT + DiRe & \bf 47.3 {\textpm} 0.1 & \bf 59.6 {\textpm} 0.2 & - \\

\bottomrule
\end{tabular}}
\end{table}

\begin{table}[hbt!]
\caption{Impact on diversity by the addition of DiRe to various DC methods on  ImageNet-1K. The diversity scores across all three metrics improve with the addition of DiRe.}
\label{ilsvrc_results_div}
\centering
\resizebox{\columnwidth}{!}{%
\begin{tabular}{lccc}
\toprule
Methods & Coverage $\uparrow$ & Similarity $\downarrow$ & Vendi $\uparrow$ \\
\toprule 
SRe$^2$L& 2.0\% &  0.82 & 4.41 \\
SRe$^2$L + DiRe&\bf 6.4\% & \bf 0.66 & \bf 5.94\\
\midrule
DWA& 2.2\% & 0.78 & 5.09 \\
DWA + DiRe&\bf 3.7\% & \bf 0.65 & \bf 5.79 \\
\midrule 
CDA & 4.1\% & 0.80 & 5.15 \\
CDA + DiRe & \bf 7.9\% & \bf 0.59 & \bf 6.18 \\
\midrule 
INFER  & 6.5\% & 0.71 & 7.45 \\
INFER + DiRe & \bf 8.3\% & \bf 0.64 & \bf 8.68 \\
\midrule 
G-VBSM & 4.1$\%$ & 0.81 & 7.87 \\
G-VBSM + DiRe & \bf 11.6$\%$ & \bf 0.67 & \bf 13.2 \\
\midrule 
DELT & 2.3$\%$ & 0.83 & 4.28 \\
DELT + DiRe & \bf 6.5$\%$ & \bf 0.65 & \bf 6.23 \\
\bottomrule
\end{tabular}}
\end{table}

\subsection{Cross-architecture study}
To evaluate the generalizability of the synthetic dataset generated by our regularizer, we compare the test set accuracies on various deep learning architectures including CNN architectures from different families, ResNet-50~\citep{he2016deep}, ResNet-101~\citep{he2016deep}, MobileNetV2~\citep{MobileNetV2_2018_CVPR}, and VGG-16~\citep{Simonyan2015VeryDC}, and transformer architecture, ViT~\citep{dosovitskiy2021an}. Comparisons of test set accuracies are reported in Tables~\ref{cifar100_cross_architecture},~\ref{tiny_cross_architecture}, and~\ref{ilsvrc_cross_architecture} for CIFAR-100, Tiny ImageNet, and ImageNet-1K, respectively. Our proposed regularizer is able to improve accuracy across various CNN and transformer architectures.

\begin{table}[hbt!]
\caption{Comparison of cross-architecture generalization performance on CIFAR-100 generated by ResNet-18.}
\label{cifar100_cross_architecture}
\begin{center}
\centering
\begin{tabular}{l|cc}
\toprule 
Target Architectures & SRe$^2$L & SRe$^2$L + DiRe \\
\toprule 
\multicolumn{3}{c}{IPC=$50$} \\
\midrule 
ResNet-50 & 52.8 {\textpm} 0.7 & \bf 63.8 {\textpm} 0.6 \\
ResNet-101 & 51.4 {\textpm} 2.6 & \bf 64.2 {\textpm} 0.3 \\
VGG-16 & 40.4 {\textpm} 1.2 & \bf 53.4 {\textpm} 0.4 \\
MobileNetV2 & 43.2 {\textpm} 0.2 & \bf 56.7 {\textpm} 0.1 \\
ViT & 16.2 {\textpm} 0.2 & \bf 35.0 {\textpm} 0.5 \\
\midrule 
\multicolumn{3}{c}{IPC=$100$} \\
\midrule 
ResNet-50 & 59.5 {\textpm} 0.5 & \bf 67.3 {\textpm} 0.2 \\
ResNet-101 & 59.2 {\textpm} 0.9 & \bf 67.7 {\textpm} 0.1 \\
VGG-16 & 51.8 {\textpm} 0.4 & \bf 62.2 {\textpm} 0.4 \\
MobileNetV2 & 54.6 {\textpm} 0.5 & \bf 64.1 {\textpm} 0.3 \\
ViT & 23.3 {\textpm} 0.4 & \bf 46.6 {\textpm} 0.9 \\
\bottomrule
\end{tabular}
\end{center}
\end{table}

\begin{table}[hbt!]
\caption{Comparison of cross-architecture generalization performance on Tiny ImageNet generated by ResNet-18.}
\label{tiny_cross_architecture}
\begin{center}
\begin{small}
\begin{tabular}{l|cc}
\toprule

Methods & VGG-16 & ViT \\
\toprule
\multicolumn{3}{c}{IPC=50} \\
\midrule  
SRe$^2$L & 32.50 {\textpm} 0.45 & 14.79 {\textpm} 0.40 \\
DWA & 32.90 {\textpm} 0.85 & 18.79 {\textpm} 0.28 \\
CDA & 33.50 {\textpm} 0.20 & 16.35 {\textpm} 0.51 \\
SRe$^2$L + DiRe & \bf 34.45 {\textpm} 0.56 & \bf 19.00 {\textpm} 0.10\\
\midrule
\multicolumn{3}{c}{IPC=100} \\
\midrule  
SRe$^2$L & 42.80 {\textpm} 0.46 & 22.95 {\textpm} 0.75 \\
DWA & 43.58 {\textpm} 0.12 & 26.46 {\textpm} 0.21 \\
CDA & 41.81 {\textpm} 0.35 & 22.65 {\textpm} 0.70 \\
SRe$^2$L + DiRe & \bf 45.21 {\textpm} 0.84 & \bf 28.20 {\textpm} 0.33\\

\bottomrule
\end{tabular}
\end{small}
\end{center}
\end{table}

\begin{table}[hbt!]
\caption{Comparison of cross-architecture generalization performance on ImageNet-1K generated by ResNet-18. Addition of DiRe is able to maintain performance gain in terms of accuracy across various architectures.}
\label{ilsvrc_cross_architecture}
\begin{center}
\begin{small}
\begin{tabular}{l|cc}
\toprule

Methods & MobileNetV2 & ShuffleNet \\
\toprule
\multicolumn{3}{c}{IPC=10} \\
\midrule  
SRe$^2$L & 15.4 {\textpm} 0.2 & 9.0 {\textpm} 0.7 \\
DWA & 29.1 {\textpm} 0.3 & \bf 11.4 {\textpm} 0.6 \\
CDA & 25.1 {\textpm} 0.5 & 8.5 {\textpm} 0.7 \\
SRe$^2$L + DiRe & \bf 30.6 {\textpm} 0.3 &  10.5 {\textpm} 0.3\\
\midrule
\multicolumn{3}{c}{IPC=50} \\
\midrule  
SRe$^2$L & 48.3 {\textpm} 0.5 & 9.0 {\textpm} 0.6 \\
DWA & 51.6 {\textpm} 0.5 & 28.5 {\textpm} 0.5 \\
CDA & 49.4 {\textpm} 0.3 & 25.8 {\textpm} 0.4 \\
SRe$^2$L + DiRe & \bf 53.0 {\textpm} 0.2 & \bf 30.0 {\textpm} 0.3 \\

\bottomrule
\end{tabular}
\end{small}
\end{center}
\end{table}

\subsection{Impact of the proposed regularizer on other condensation methods}
We apply DiRe to MTT and DM to demonstrate its effectiveness on DC algorithms that are not decoupling-based. Table~\ref{mtt_analysis} shows the accuracy results obtained on a ConvNet network on CIFAR-10 and CIFAR-100 for various IPC settings with MTT algorithm. Results for DM are given in the supplementary document.

\begin{table}[hbt!]
\caption{Comparison of accuracies on ConvNet architecture with MTT Dataset Condensation method. DiRe is able to improve the performance of the non-decoupled algorithm across various IPC settings.}
\label{mtt_analysis}
\begin{center}
\centering
\begin{tabular}{lcc}
\toprule 
IPC & MTT & MTT + DiRe \\
\midrule 
\multicolumn{3}{c}{CIFAR-10} \\
\midrule
1 & 46.3 {\textpm} 0.8 & \bf 49.1 {\textpm} 0.2\\
10 & 65.3 {\textpm} 0.7 & \bf 67.4 {\textpm} 0.3 \\
50 & 71.6 {\textpm} 0.2 & \bf 72.4 {\textpm} 0.1 \\
\midrule 
\multicolumn{3}{c}{CIFAR-100} \\
\midrule 
1 & 24.3 {\textpm} 0.3 & \bf 27.6 {\textpm} 0.7 \\
10 & 40.1 {\textpm} 0.4 & \bf 42.0 {\textpm} 0.2 \\
50 & 47.7 {\textpm} 0.2 & \bf 48.9 {\textpm} 0.3 \\
\bottomrule
\end{tabular}
\end{center}
\end{table}

\subsection{Ablation study on impact of various components of DiRe}
We analyze the impact of individual components in the DiRe regularizer and their combinations. We compute diversity metrics for a synthetic dataset using all seven possible combinations. Table~\ref{tab:heatmap} compares the normalized impact of various components of DiRe and their combinations on test set accuracy and different diversity metrics for the Tiny ImageNet dataset with IPC = $10$. As can be seen, CD mainly affects cosine similarity, CDM primarily influences the Vendi score, and EDM has the strongest impact on the coverage score. Combining all three components together results in balanced performance across all diversity metrics.
\begin{table}[hbt!]
\caption{Normalized impact of various components of DiRe and their combinations on accuracy and different diversity metrics for Tiny ImageNet on IPC=$10$ setting. Ven, Sim, and Cov stand for Vendi Score, Cosine Similarity, and Coverage, respectively. }
\label{tab:heatmap}
\centering
\resizebox{\columnwidth}{!}{%
\begin{tabular}{lcccc}
\toprule
Methods & Ven$\uparrow$ & Sim$\downarrow$ & Cov$\uparrow$& Acc\\
\toprule 
CD & 0.50 & 0.00 & 0.41 & 33.8 {\textpm} 0.4\\
CDM & 0.74 & 0.30 & 0.46 & 33.8 {\textpm} 0.2\\
EDM & 0.00 & 0.98 & 1.00 & 34.2 {\textpm} 0.3\\
CD + CDM & 0.48 & 0.17 & 0.00 & 34.1 {\textpm} 0.6 \\
CD + EDM & 0.40 & 0.93 & 0.01 & 33.6 {\textpm} 0.1\\
CDM + EDM & 0.36 & 1.00 & 0.92 & 33.6 {\textpm} 0.5 \\
CD + CDM + EDM & 1.00 & 0.07 & 0.80 & 34.7 {\textpm} 0.3\\
\bottomrule
\end{tabular}}
\end{table}
\subsection{Timing analysis}
\label{timing_analysis}
Table~\ref{timing_analysis_result} compares the time taken by various methods to synthesize IPC=$50$ images from CIFAR-100. 
The synthesis time with DiRe is $\approx 17\%$ higher than SRe$^2$L and $\approx 5\%$ higher than DWA, which we believe is modest compared to the gains in generalization and diversity. A comparative analysis of compute requirements is provided in the supplementary material.

\begin{table}[hbt!]
\caption{Comparison of total time taken for synthesizing IPC=50 images from the CIFAR-100 by various methods.}
\label{timing_analysis_result}
\begin{center}
\begin{tabular}{l|c}
\toprule
Methods & Time (in seconds) \\
\midrule
SRe$^2$L & 1342.6 \\
DWA & 1493.1 \\
SRe$^2$L + DiRe & 1579.9 \\
\bottomrule
\end{tabular}
\end{center}
\end{table}

\subsection{Calculation of Coverage \& Vendi Score through CLIP embeddings}
For completeness and to avoid any bias introduced by a specific backbone, we further evaluate coverage and Vendi Score using embeddings extracted from a pretrained CLIP model~\citep{radford2021learning} (ViT-B/32, trained on the LAION-2B dataset). Unlike the ResNet-18 features used in our primary experiments, CLIP embeddings are obtained via a vision–language pretraining paradigm, which has been shown to produce semantically aligned and domain-robust representations. Comparative analysis is provided in Table~\ref{clip_analysis}.
\begin{table}[hbt!]
\caption{Comparison of coverage values obtained by ResNet-18 and CLIP pre-trained models. Results are obtained on Tiny ImageNet with synthetic data generated through a ResNet-18 architecture.}
\label{clip_analysis}
\begin{center}
\resizebox{\columnwidth}{!}{%
\begin{tabular}{l|cc|cc}
\toprule
& \multicolumn{2}{c|}{Coverage} & \multicolumn{2}{c}{Vendi Score} \\
\midrule
Methods & ResNet-18 & CLIP & ResNet-18 & CLIP\\
\midrule
SRe$^2$L & 0.301 & 0.064 & 3.07 & 2.08\\
SRe$^2$L + DiRe & \bf 0.453 & \bf 0.088 & \bf 3.22 & \bf 2.42 \\
\bottomrule
\end{tabular}}
\end{center}
\end{table}

This additional evaluation confirms that our conclusions remain consistent even when employing a feature extractor with a substantially different training objective and representational bias.
\section{Discussion on optimization-free algorithms}
As mentioned in Sec.~\ref{sec:introduction}, our proposed regularizer is not applicable to optimization-free algorithms such as RDED. However, for completeness, we compare the performance of SRe$^2$L + DiRe against RDED in terms of various diversity metrics on ImageNet-1K. As RDED and SRe$^2$L employ different hyperparameters, results presented in Table~\ref{rded_comparison} are obtained by SRe$^2$L + DiRe with the same hyperparameter settings as RDED. We observe that SRe$^2$L + DiRe performs better than RDED across all three diversity metrics. 

\begin{table}[hbt!]
\caption{Comparative analysis of diversity metrics for synthetic data generated through RDED and SRe$^2$L + DiRe from ImageNet-1K. SRe$^2$L + DiRe outperforms RDED across the three diversity metrics. }
\centering
\begin{tabular}{l|ccc}
    \toprule
    Methods  & Coverage$\uparrow$ & Similarity$\downarrow$ & Vendi$\uparrow$ \\
    \midrule
    RDED & 0.40 & 0.71 & 4.54 \\
    SRe$^2$L + DiRe & \bf 0.45 & \bf 0.66 & \bf 6.17 \\
    \bottomrule
\end{tabular}
\label{rded_comparison}
\end{table}
\section{Conclusion and Future DiRections}
Despite its importance, diversity has not been treated as a first-class citizen in existing DC works. This paper emphasizes the significance of diversity in dataset condensation and introduces an intuitive yet powerful diversity regularizer \textbf{DiRe}, which can be added off-the-shelf to existing optimization-based dataset condensation algorithms with a separate synthesis stage. To the best of our knowledge, this is the first work to study diversity in dataset condensation in a quantitative manner. Through extensive experiments, we demonstrate that DiRe can improve the accuracy of the SOTA dataset condensation methods while achieving higher diversity across various diversity metrics. We also demonstrate that the improved diversity helps to achieve better cross-architecture generalization on various deep learning architectures, including transformers.

Some avenues for future work include:  making the Euclidean distance computation in DiRe more efficient using coresets~\citep{har2004coresets}, applying DiRe in the field of generative modeling, and using functions such as the Determinantal Point Process objective~\citep{celis2018fair} or other submodular functions~\citep{bilmes2022submodularity} instead of cosine similarity for devising new diversity regularizers.

\clearpage
\section*{Acknowledgements}
Most of this work was done while Aravind Reddy was at the Department of Artificial Intelligence, Indian Institute of Technology Hyderabad.
{
    \small
    \bibliographystyle{ieeenat_fullname}
    \bibliography{ref}

@STRING{aaai	= {AAAI} }

@STRING{cvpr	= {CVPR} }

@STRING{iccv	= {ICCV} }

@STRING{iclr	= {ICLR} }

@STRING{icml	= {ICML} }

@STRING{ijcv	= {IJCV} }

@STRING{neurips	= {NeurIPS} }

@STRING{stoc	= {STOC} }

@STRING{tmlr	= {TMLR} }

@STRING{wacv	= {WACV} }

@Article{	  wang2018dataset,
  title		= {Dataset distillation},
  author	= {Wang, Tongzhou and Zhu, Jun-Yan and Torralba, Antonio and
		  Efros, Alexei A},
  journal	= {arXiv:1811.10959},
  year		= {2018}
}

@InProceedings{	  yinsqueeze,
  title		= {Squeeze, Recover and Relabel: Dataset Condensation at
		  ImageNet Scale From A New Perspective},
  author	= {Zeyuan Yin and Eric Xing and Zhiqiang Shen},
  booktitle	= neurips,
  year		= {2023}
}

@InProceedings{	  du2024diversity,
  title		= {Diversity-Driven Synthesis: Enhancing Dataset Distillation
		  through Directed Weight Adjustment},
  author	= {Jiawei Du and Xin Zhang and Juncheng Hu and Wenxin Huang
		  and Joey Tianyi Zhou},
  booktitle	= neurips,
  year		= {2024},
  url		= {https://openreview.net/forum?id=uwSaDHLlYc}
}

@InProceedings{	  delt_2025_cvpr,
  author	= {Shen, Zhiqiang and Sherif, Ammar and Yin, Zeyuan and Shao,
		  Shitong},
  title		= {DELT: A Simple Diversity-driven EarlyLate Training for
		  Dataset Distillation},
  booktitle	= cvpr,
  year		= {2025}
}

@Article{	  yin2024cda,
  title		= {Dataset Distillation via Curriculum Data Synthesis in
		  Large Data Era},
  author	= {Zeyuan Yin and Zhiqiang Shen},
  journal	= tmlr,
  year		= {2024}
}

@InProceedings{	  he2016deep,
  title		= {Deep residual learning for image recognition},
  author	= {He, Kaiming and Zhang, Xiangyu and Ren, Shaoqing and Sun,
		  Jian},
  booktitle	= cvpr,
  year		= {2016}
}

@Article{	  russakovsky2015imagenet,
  title		= {ImageNet Large Scale Visual Recognition Challenge},
  author	= {Russakovsky, Olga and Deng, Jia and Su, Hao and Krause,
		  Jonathan and Satheesh, Sanjeev and Ma, Sean and Huang,
		  Zhiheng and Karpathy, Andrej and Khosla, Aditya and
		  Bernstein, Michael and Berg, Alexander C. and {Fei-Fei},
		  Li},
  year		= 2015,
  journal	= ijcv
}

@Article{	  lee2021vision,
  title		= {Vision transformer for small-size datasets},
  author	= {Lee, Seung Hoon and Lee, Seunghyun and Song, Byung Cheol},
  journal	= {arXiv:2112.13492},
  year		= {2021}
}

@Article{	  krizhevsky2009learning,
  title		= {Learning Multiple Layers of Features from Tiny Images},
  author	= {Krizhevsky, Alex},
  journal	= {Master's thesis, University of Toronto},
  pages		= {32--33},
  year		= {2009}
}

@Misc{		  le2015tinyiv,
  title		= {Tiny ImageNet Visual Recognition Challenge},
  author	= {Ya Le and Xuan Yang},
  year		= {2015},
  url		= {http://vision.stanford.edu/teaching/cs231n/reports/2015/pdfs/yle_project.pdf}
}

@InProceedings{	  xia2023moderate,
  title		= {Moderate Coreset: A Universal Method of Data Selection for
		  Real-world Data-efficient Deep Learning},
  author	= {Xiaobo Xia and Jiale Liu and Jun Yu and Xu Shen and Bo Han
		  and Tongliang Liu},
  booktitle	= iclr,
  year		= {2023}
}

@InProceedings{	  naeem2020reliable,
  title		= {Reliable Fidelity and Diversity Metrics for Generative
		  Models},
  author	= {Naeem, Muhammad Ferjad and Oh, Seong Joon and Uh,
		  Youngjung and Choi, Yunjey and Yoo, Jaejun},
  booktitle	= icml,
  year		= {2020},
  url		= {https://proceedings.mlr.press/v119/naeem20a.html}
}

@Article{	  friedman2023the,
  title		= {The Vendi Score: A Diversity Evaluation Metric for Machine
		  Learning},
  author	= {Dan Friedman and Adji Bousso Dieng},
  journal	= tmlr,
  issn		= {2835-8856},
  year		= {2023},
  url		= {https://openreview.net/forum?id=g97OHbQyk1},
  note		= {}
}

@Article{	  diversityml2019,
  author	= {Gong, Zhiqiang and Zhong, Ping and Hu, Weidong},
  journal	= {IEEE Access},
  title		= {Diversity in Machine Learning},
  year		= {2019}
}

@InProceedings{	  zhao2024position,
  title		= {Position: Measure Dataset Diversity, Don't Just Claim It},
  author	= {Dora Zhao and Jerone Andrews and Orestis Papakyriakopoulos
		  and Alice Xiang},
  booktitle	= icml,
  year		= {2024}
}

@InProceedings{	  sun2024diversity,
  title		= {On the diversity and realism of distilled dataset: An
		  efficient dataset distillation paradigm},
  author	= {Sun, Peng and Shi, Bei and Yu, Daiwei and Lin, Tao},
  booktitle	= cvpr,
  year		= {2024}
}

@InProceedings{	  zhao2021dataset,
  title		= {Dataset Condensation with Gradient Matching},
  author	= {Bo Zhao and Konda Reddy Mopuri and Hakan Bilen},
  booktitle	= iclr,
  year		= {2021}
}

@InProceedings{	  lee2024selmatch,
  title		= {SelMatch: Effectively Scaling Up Dataset Distillation via
		  Selection-Based Initialization and Partial Updates by
		  Trajectory Matching},
  author	= {Yongmin Lee and Hye Won Chung},
  booktitle	= icml,
  year		= {2024},
  url		= {https://openreview.net/forum?id=pTFud6SetK}
}

@InProceedings{	  yang2024what,
  title		= {What is Dataset Distillation Learning?},
  author	= {William Yang and Ye Zhu and Zhiwei Deng and Olga
		  Russakovsky},
  booktitle	= icml,
  year		= {2024}
}

@InProceedings{	  cui2024ameliorate,
  title		= {Ameliorate Spurious Correlations in Dataset Condensation},
  author	= {Justin Cui and Ruochen Wang and Yuanhao Xiong and Cho-Jui
		  Hsieh},
  booktitle	= icml,
  year		= {2024},
  url		= {https://openreview.net/forum?id=RbnojVv4HK}
}

@InProceedings{	  qin2024a,
  title		= {A Label is Worth A Thousand Images in Dataset
		  Distillation},
  author	= {Tian Qin and Zhiwei Deng and David Alvarez-Melis},
  booktitle	= neurips,
  year		= {2024},
  url		= {https://openreview.net/forum?id=oNMnR0NJ2e}
}

@InProceedings{	  ding2024condtsf,
  title		= {Cond{TSF}: One-line Plugin of Dataset Condensation for
		  Time Series Forecasting},
  author	= {Jianrong Ding and Zhanyu Liu and Guanjie Zheng and Haiming
		  Jin and Linghe Kong},
  booktitle	= neurips,
  year		= {2024},
  url		= {https://openreview.net/forum?id=L1jajNWON5}
}

@InProceedings{	  qi2024fetch,
  title		= {Fetch and Forge: Efficient Dataset Condensation for Object
		  Detection},
  author	= {Ding Qi and Jian Li and Jinlong Peng and Bo Zhao and
		  Shuguang Dou and Jialin Li and Jiangning Zhang and Yabiao
		  Wang and Chengjie Wang and Cairong Zhao},
  booktitle	= neurips,
  year		= {2024},
  url		= {https://openreview.net/forum?id=m8MElyzuwp}
}

@InProceedings{	  shao2024elucidating,
  title		= {Elucidating the Design Space of Dataset Condensation},
  author	= {Shitong Shao and Zikai Zhou and Huanran Chen and Zhiqiang
		  Shen},
  booktitle	= neurips,
  year		= {2024},
  url		= {https://openreview.net/forum?id=az1SLLsmdR}
}

@InProceedings{	  dosovitskiy2021an,
  title		= {An Image is Worth 16x16 Words: Transformers for Image
		  Recognition at Scale},
  author	= {Alexey Dosovitskiy and Lucas Beyer and Alexander
		  Kolesnikov and Dirk Weissenborn and Xiaohua Zhai and Thomas
		  Unterthiner and Mostafa Dehghani and Matthias Minderer and
		  Georg Heigold and Sylvain Gelly and Jakob Uszkoreit and
		  Neil Houlsby},
  booktitle	= iclr,
  year		= {2021},
  url		= {https://openreview.net/forum?id=YicbFdNTTy}
}

@InProceedings{	  mobilenetv2_2018_cvpr,
  author	= {Sandler, Mark and Howard, Andrew and Zhu, Menglong and
		  Zhmoginov, Andrey and Chen, Liang-Chieh},
  title		= {MobileNetV2: Inverted Residuals and Linear Bottlenecks},
  booktitle	= cvpr,
  month		= {June},
  year		= {2018}
}

@InProceedings{	  simonyan2015verydc,
  title		= {Very Deep Convolutional Networks for Large-Scale Image
		  Recognition},
  author	= {Karen Simonyan and Andrew Zisserman},
  booktitle	= iclr,
  year		= {2015}
}

@Article{	  sachdeva2023data,
  title		= {Data Distillation: A Survey},
  author	= {Noveen Sachdeva and Julian McAuley},
  journal	= tmlr,
  issn		= {2835-8856},
  year		= {2023},
  url		= {https://openreview.net/forum?id=lmXMXP74TO}
}

@InProceedings{	  deng2022remember,
  title		= {Remember the Past: Distilling Datasets into Addressable
		  Memories for Neural Networks},
  author	= {Zhiwei Deng and Olga Russakovsky},
  booktitle	= neurips,
  editor	= {Alice H. Oh and Alekh Agarwal and Danielle Belgrave and
		  Kyunghyun Cho},
  year		= {2022},
  url		= {https://openreview.net/forum?id=RYZyj_wwgfa}
}

@InProceedings{	  gu2024summarizing,
  title		= {Summarizing Stream Data for Memory-Constrained Online
		  Continual Learning},
  author	= {Gu, Jianyang and Wang, Kai and Jiang, Wei and You, Yang},
  booktitle	= aaai,
  year		= {2024},
  url		= {https://ojs.aaai.org/index.php/AAAI/article/view/29111}
}

@InProceedings{	  dong2022privacy,
  title		= {Privacy for Free: How does Dataset Condensation Help
		  Privacy?},
  author	= {Dong, Tian and Zhao, Bo and Lyu, Lingjuan},
  booktitle	= icml,
  year		= {2022},
  url		= {https://proceedings.mlr.press/v162/dong22c.html}
}

@InProceedings{	  chung2024rethinking,
  title		= {Rethinking Backdoor Attacks on Dataset Distillation: A
		  Kernel Method Perspective},
  author	= {Ming-Yu Chung and Sheng-Yen Chou and Chia-Mu Yu and Pin-Yu
		  Chen and Sy-Yen Kuo and Tsung-Yi Ho},
  booktitle	= iclr,
  year		= {2024},
  url		= {https://openreview.net/forum?id=iCNOK45Csv}
}

@InProceedings{	  loo2024understanding,
  title		= {Understanding Reconstruction Attacks with the Neural
		  Tangent Kernel and Dataset Distillation},
  author	= {Noel Loo and Ramin Hasani and Mathias Lechner and
		  Alexander Amini and Daniela Rus},
  booktitle	= iclr,
  year		= {2024},
  url		= {https://openreview.net/forum?id=VoLDkQ6yR3}
}

@InProceedings{	  such2020generative,
  title		= {Generative Teaching Networks: Accelerating Neural
		  Architecture Search by Learning to Generate Synthetic
		  Training Data},
  author	= {Such, Felipe Petroski and Rawal, Aditya and Lehman, Joel
		  and Stanley, Kenneth and Clune, Jeffrey},
  booktitle	= icml,
  year		= {2020}
}

@InProceedings{	  huang2024overcoming,
  title		= {Overcoming Data and Model heterogeneities in Decentralized
		  Federated Learning via Synthetic Anchors},
  author	= {Chun-Yin Huang and Kartik Srinivas and Xin Zhang and
		  Xiaoxiao Li},
  booktitle	= icml,
  year		= {2024},
  url		= {https://openreview.net/forum?id=mNzkumTSVL}
}

@InProceedings{	  xiong2023feddm,
  title		= {Feddm: Iterative distribution matching for
		  communication-efficient federated learning},
  author	= {Xiong, Yuanhao and Wang, Ruochen and Cheng, Minhao and Yu,
		  Felix and Hsieh, Cho-Jui},
  booktitle	= cvpr,
  year		= {2023}
}

@InProceedings{	  jain2023efficient,
  title		= {Efficient Data Subset Selection to Generalize Training
		  Across Models: Transductive and Inductive Networks},
  author	= {Eeshaan Jain and Tushar Nandy and Gaurav Aggarwal and
		  Ashish V. Tendulkar and Rishabh K Iyer and Abir De},
  booktitle	= neurips,
  year		= {2023},
  url		= {https://openreview.net/forum?id=q3fCWoC9l0}
}

@InProceedings{	  kai2015submodularity,
  title		= {Submodularity in Data Subset Selection and Active
		  Learning},
  author	= {Wei, Kai and Iyer, Rishabh and Bilmes, Jeff},
  booktitle	= icml,
  year		= {2015}
}

@InProceedings{	  yang2023towards,
  title		= {Towards Sustainable Learning: Coresets for Data-efficient
		  Deep Learning},
  author	= {Yang, Yu and Kang, Hao and Mirzasoleiman, Baharan},
  booktitle	= icml,
  year		= {2023}
}

@InProceedings{	  yang2023dataset,
  title		= {Dataset Pruning: Reducing Training Data by Examining
		  Generalization Influence},
  author	= {Shuo Yang and Zeke Xie and Hanyu Peng and Min Xu and
		  Mingming Sun and Ping Li},
  booktitle	= iclr,
  year		= {2023}
}

@InProceedings{	  zhang2023selectivity,
  title		= {Selectivity Drives Productivity: Efficient Dataset Pruning
		  for Enhanced Transfer Learning},
  author	= {Yihua Zhang and Yimeng Zhang and Aochuan Chen and Jinghan
		  Jia and Jiancheng Liu and Gaowen Liu and Mingyi Hong and
		  Shiyu Chang and Sijia Liu},
  booktitle	= neurips,
  year		= {2023}
}

@InProceedings{	  zhang2024spanning,
  title		= {Spanning training progress: Temporal dual-depth scoring
		  (tdds) for enhanced dataset pruning},
  author	= {Zhang, Xin and Du, Jiawei and Li, Yunsong and Xie, Weiying
		  and Zhou, Joey Tianyi},
  booktitle	= cvpr,
  year		= {2024}
}

@InProceedings{	  har2004coresets,
  title		= {On coresets for k-means and k-median clustering},
  author	= {Har-Peled, Sariel and Mazumdar, Soham},
  booktitle	= stoc,
  year		= {2004}
}

@InProceedings{	  zhao2021siamese,
  title		= {Dataset Condensation with Differentiable Siamese
		  Augmentation},
  author	= {Zhao, Bo and Bilen, Hakan},
  booktitle	= icml,
  year		= {2021}
}

@InProceedings{	  cazenavette2022cvpr,
  title		= {Dataset Distillation by Matching Training Trajectories},
  author	= {George Cazenavette and Tongzhou Wang and Antonio Torralba
		  and Alexei A. Efros and Jun-Yan Zhu},
  booktitle	= cvpr,
  year		= {2022}
}

@InProceedings{	  guo2024towards,
  title		= {Towards Lossless Dataset Distillation via
		  Difficulty-Aligned Trajectory Matching},
  author	= {Ziyao Guo and Kai Wang and George Cazenavette and Hui Li
		  and Kaipeng Zhang and Yang You},
  booktitle	= iclr,
  year		= {2024},
  url		= {https://openreview.net/forum?id=rTBL8OhdhH}
}

@InProceedings{	  zhao2023distribution,
  title		= {Dataset condensation with distribution matching},
  author	= {Zhao, Bo and Bilen, Hakan},
  booktitle	= wacv,
  year		= {2023}
}

@InProceedings{	  zhao2023improved,
  title		= {Improved distribution matching for dataset condensation},
  author	= {Zhao, Ganlong and Li, Guanbin and Qin, Yipeng and Yu,
		  Yizhou},
  booktitle	= cvpr,
  year		= {2023}
}

@InProceedings{	  van2021benchmarking,
  title		= {Benchmarking representation learning for natural world
		  image collections},
  author	= {Van Horn, Grant and Cole, Elijah and Beery, Sara and
		  Wilber, Kimberly and Belongie, Serge and Mac Aodha, Oisin},
  booktitle	= cvpr,
  year		= {2021}
}

@InProceedings{	  yang2021just,
  title		= {Just ask: Learning to answer questions from millions of
		  narrated videos},
  author	= {Yang, Antoine and Miech, Antoine and Sivic, Josef and
		  Laptev, Ivan and Schmid, Cordelia},
  booktitle	= iccv,
  year		= {2021}
}

@InProceedings{	  zhou2022dataset,
  title		= {Dataset Distillation using Neural Feature Regression},
  author	= {Yongchao Zhou and Ehsan Nezhadarya and Jimmy Ba},
  booktitle	= neurips,
  year		= {2022}
}

@InProceedings{	  shao2024generalized,
  author	= {Shao, Shitong and Yin, Zeyuan and Zhou, Muxin and Zhang,
		  Xindong and Shen, Zhiqiang},
  title		= {Generalized Large-Scale Data Condensation via Various
		  Backbone and Statistical Matching},
  booktitle	= cvpr,
  month		= {June},
  year		= {2024}
}

@InProceedings{	  celis2018fair,
  title		= {Fair and Diverse {DPP}-Based Data Summarization},
  author	= {Celis, Elisa and Keswani, Vijay and Straszak, Damian and
		  Deshpande, Amit and Kathuria, Tarun and Vishnoi, Nisheeth},
  booktitle	= icml,
  year		= {2018}
}

@Article{	  bilmes2022submodularity,
  title		= {Submodularity in machine learning and artificial
		  intelligence},
  author	= {Bilmes, Jeff},
  journal	= {arXiv:2202.00132},
  year		= {2022}
}

@InProceedings{	  zhang2025breaking,
  title		= {Breaking Class Barriers: Efficient Dataset Distillation
		  via Inter-Class Feature Compensator},
  author	= {Xin Zhang and Jiawei Du and Ping Liu and Joey Tianyi
		  Zhou},
  booktitle	= iclr,
  year		= {2025}
}

@InProceedings{	  radford2021learning,
  title		= {Learning Transferable Visual Models From Natural Language
		  Supervision},
  author	= {Radford, Alec and Kim, Jong Wook and Hallacy, Chris and
		  Ramesh, Aditya and Goh, Gabriel and Agarwal, Sandhini and
		  Sastry, Girish and Askell, Amanda and Mishkin, Pamila and
		  Clark, Jack and Krueger, Gretchen and Sutskever, Ilya},
  booktitle	= icml,
  year		= {2021}
}
}
\clearpage

\onecolumn
\appendix
\section{Visualization of synthetic images}
\begin{figure*}[ht!]
\begin{center}
\centerline{\includegraphics[scale=0.2]{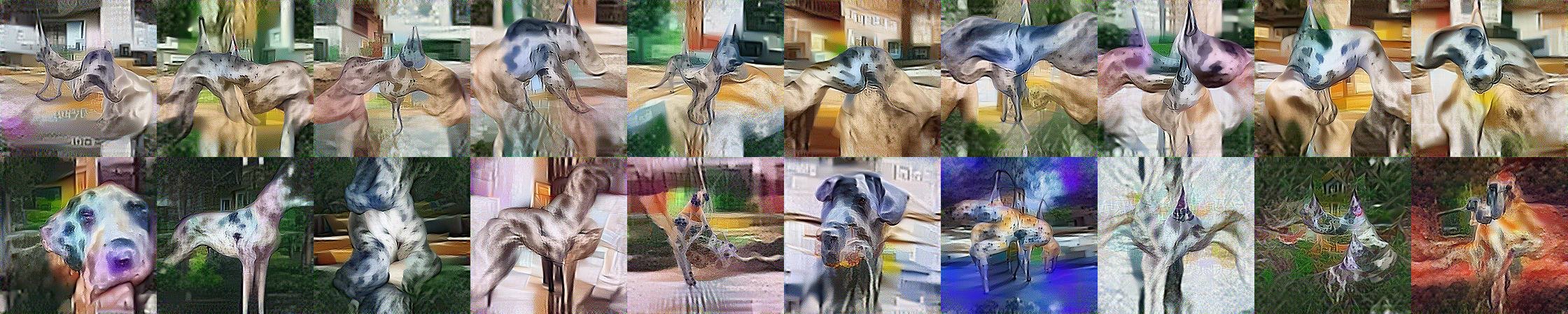}}
\caption{Visualization of synthetic images belonging to the `Great Dane' class from the ImageNet-1K dataset. The top row images are condensed through SRe$^2$L and the bottom row images are condensed through SRe$^2$L + DiRe. }
\label{ilsvrc_visualization1}
\end{center}
\end{figure*}

\begin{figure*}[ht!]
\begin{center}
\centerline{\includegraphics[scale=0.2]{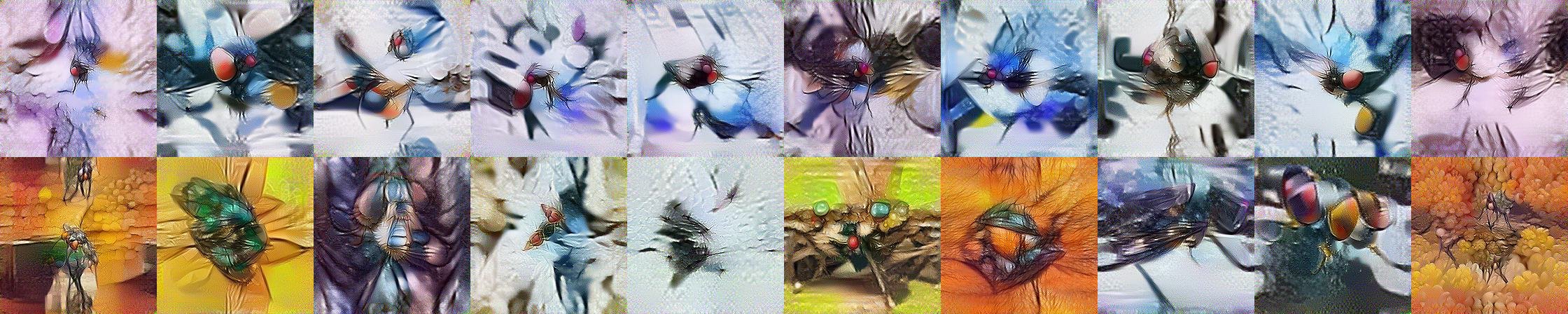}}
\caption{Visualization of synthetic images belonging to the `Fly' class from the ImageNet-1K dataset. The top row images are condensed through SRe$^2$L and the bottom row images are condensed through SRe$^2$L + DiRe. }
\label{ilsvrc_visualization2}
\end{center}
\end{figure*}

\begin{figure*}[ht!]
\begin{center}
\centerline{\includegraphics[scale=0.2]{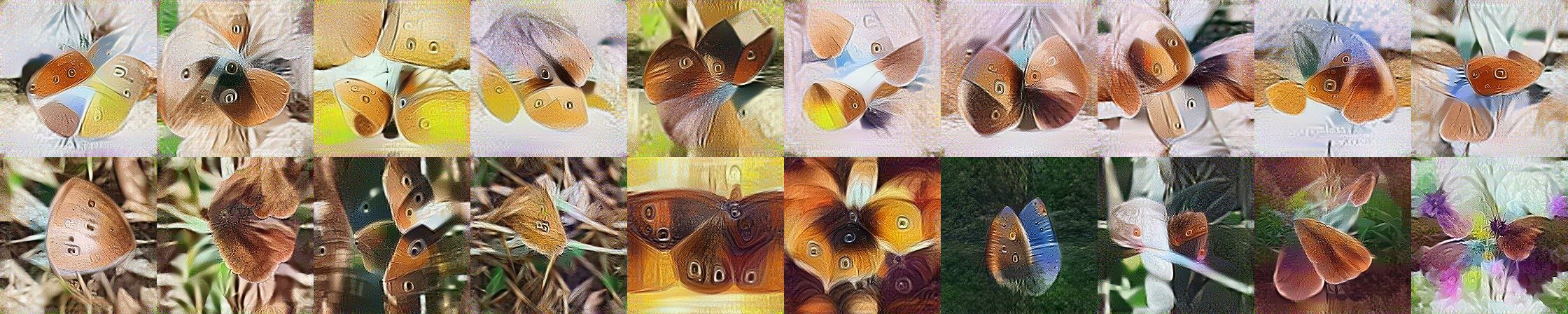}}
\caption{Visualization of synthetic images belonging to the `Ringlet Butterfly' class from the ImageNet-1K dataset. The top row images are condensed through SRe$^2$L and the bottom row images are condensed through SRe$^2$L + DiRe. }
\label{ilsvrc_visualization3}
\end{center}
\end{figure*}

\begin{figure*}[hbt!]
\begin{center}
\centerline{\includegraphics[scale=0.2]{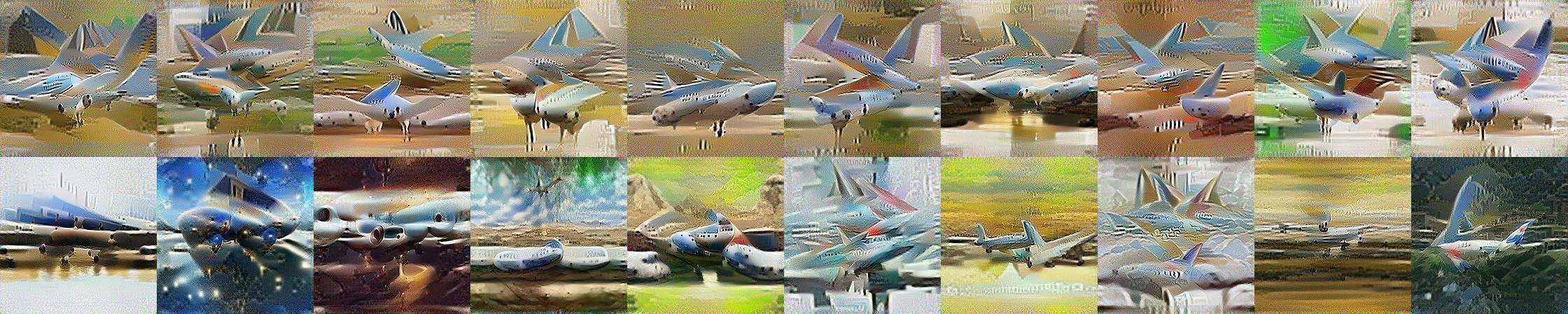}}
\caption{Visualization of synthetic images belonging to the `Airliner' class from the ImageNet-1K dataset. The top row images are condensed through SRe$^2$L and the bottom row images are condensed through SRe$^2$L + DiRe. }
\label{ilsvrc_visualization4}
\end{center}
\end{figure*}

\begin{figure*}[hbt!]
\begin{center}
\centerline{\includegraphics[scale=0.25]{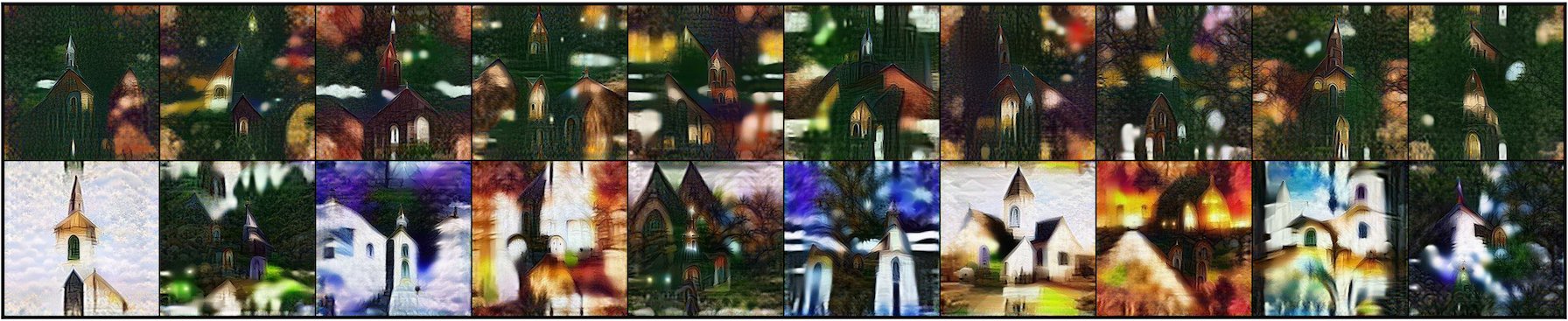}}
\caption{Visualization of synthetic images belonging to the `church' class from the ImageNet-1K dataset. The top row images are condensed through SRe$^2$L and the bottom row images are condensed through SRe$^2$L + DiRe. }
\label{ilsvrc_visualization5}
\end{center}
\end{figure*}

\begin{figure*}[hbt!]
\begin{center}
\centerline{\includegraphics[scale=0.2]{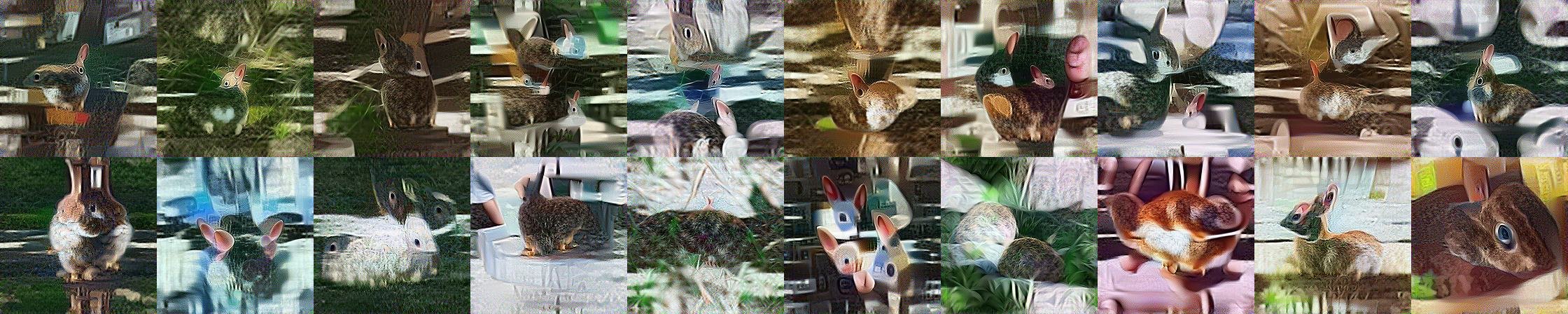}}
\caption{Visualization of synthetic images belonging to the `Rabbit' class from the ImageNet-1K dataset. The top row images are condensed through SRe$^2$L, and the bottom row images are condensed through SRe$^2$L + DiRe.}
\label{ilsvrc_visualization6}
\end{center}
\end{figure*}

Figures~\ref{ilsvrc_visualization1},~\ref{ilsvrc_visualization2},~\ref{ilsvrc_visualization3},~\ref{ilsvrc_visualization4},~\ref{ilsvrc_visualization5}, and~\ref{ilsvrc_visualization6} show visualization of synthetic images generated through SRe$^2$L and SRe$^2$L + DiRe methods. The diversity introduced by addition of DiRe is clearly visible.

\begin{figure*}[hbt!]
\begin{center}
\centerline{\includegraphics[scale=0.25]{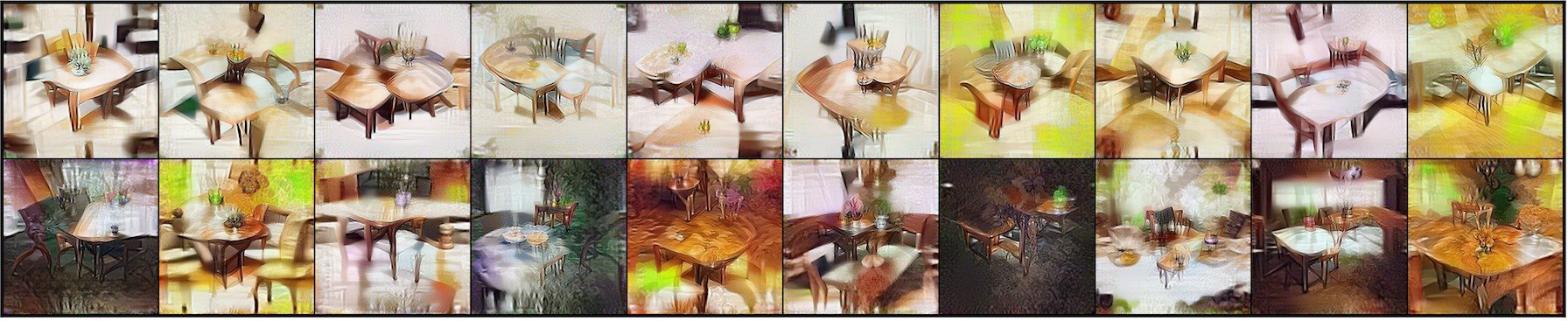}}
\caption{Visualization of synthetic images belonging to the `Dining Table' class from the ImageNet-1K dataset. The top row images are condensed through SRe$^2$L and the bottom row images are condensed through SRe$^2$L + DiRe. }
\end{center}
\end{figure*}

\newpage
\section{Intuitive motivation for \textbf{DiRe}}

In this section, we discuss some intuitive motivation for why DiRe, comprising of cosine diversity, cosine distribution matching, and Euclidean distribution matching leads to improved diversity in DC.

\subsection{Notation}

Let:
\begin{itemize}
  \item $\mathcal{D}_{\text{real}} = \{x_i, y_i\}_{i=1}^N$: original dataset
  \item $\mathcal{D}_{\text{syn}} = \{\tilde{x}_j, \tilde{y}_j\}_{j=1}^M$: synthetic dataset, with $M \ll N$
  \item $h_\theta(\cdot)$: pretrained embedding function (e.g., ResNet avgpool output)
  \item $E_{\text{real}}, E_{\text{syn}}$: embedding matrices for real and synthetic data
\end{itemize}

Goal: Ensure $\mathcal{D}_{\text{syn}}$ generalizes similarly to $\mathcal{D}_{\text{real}}$ for training neural networks.

\subsection{Cosine Diversity Loss}

Let $E^c_{\text{syn}}$ denote synthetic embeddings of class $c$.

Define the cosine diversity loss:
\[
\mathcal{L}_{\text{cos-div}} = \sum_{i < j} \cos(E_i^c, E_j^c)
\]

\begin{itemize}
    \item The Cosine similarity of two orthogonal vectors is zero. 
    \item Minimization of the sum of pairwise cosine similarities results in vectors being mutually orthogonal. 
    \item Thus, it ensures vectors are geometrically spread out, resulting in higher diversity. 
\end{itemize}

More formally, we can relate the minimization of pairwise cosine similarity to the maximization of the determinant of a corresponding Gram matrix. The determinant of the Gram matrix is related to the volume spanned by the vectors in the feature space. A larger determinant indicates that the vectors are more spread out, corresponding to higher diversity. 

If the vectors are highly similar (high cosine similarities), they are nearly collinear and the Gram matrix becomes close to singular (low determinant). Conversely, if the vectors are orthogonal (cosine similarities close to 0), the Gram matrix approaches a diagonal matrix with a determinant of 1, indicating a maximal spread for the vectors.

\subsection{Distribution Matching Losses}

Define:

\begin{align*}
\mathcal{L}_{\text{cos-dm}} &= \sum_{i,j} \left(1 - \cos(E^{c}_{\text{syn},i}, E^{c}_{\text{real},j}) \right) \\
\mathcal{L}_{\text{euc-dm}} &= \sum_{i,j} \| E^{c}_{\text{syn},i} - E^{c}_{\text{real},j} \|_2
\end{align*}

Minimizing both losses jointly aligns synthetic data with real data in:
\begin{itemize}
  \item \textbf{Direction} (via cosine similarity)
  \item \textbf{Magnitude} (via Euclidean distance)
\end{itemize}

This encourages synthetic data to approximate the local geometry of the real data manifold.

\subsection{Conclusion}

Thus DiRe promotes diversity and alignment in embedding space. This:
\begin{itemize}
  \item Maximizes class-wise dispersion.
  \item Aligns synthetic samples with the data manifold.
\end{itemize}

\newpage
\section{Results for Distribution Matching (DM) method}
Comparison of test set accuracies on CIFAR-10 and CIFAR-100 with a ConvNet architecture for DM  and DM + DiRe methods is given in Table~\ref{dm_analysis}. Addition of DiRe is able to enhance accuracy values obtained by DM for various IPC values considered.
\begin{table}[hbt!]
\caption{Comparison of accuracies on ConvNet architecture with DM Dataset Condensation method.}
\label{dm_analysis}
\begin{center}
\centering
\begin{tabular}{lcc}
\toprule 
IPC & DM & DM + DiRe \\
\midrule
\multicolumn{3}{c}{CIFAR-10} \\
\midrule 
$10$ & 48.9 {\textpm} 0.6 & \bf 51.6 {\textpm} 0.3 \\
$50$ & 63.0 {\textpm} 0.4 & \bf 64.5 {\textpm} 0.1 \\
\midrule
\multicolumn{3}{c}{CIFAR-100} \\
\midrule 
$10$ & 29.7 {\textpm} 0.3 & \bf 31.8 {\textpm} 0.5 \\
$50$ & 43.6 {\textpm} 0.4 & \bf 44.9 {\textpm} 0.3 \\
\bottomrule
\end{tabular}
\end{center}
\end{table}

\section{Compute analysis}
We carry out synthesis in a single 32 GB V100 GPU. The timing and memory requirement comparison for synthesizing IPC=$10$ images from ImageNet-1K dataset is provided in Table~\ref{timing_analysis_result_ilsvrc}. It may be noted that, similar to SRe$^2$L, all the computation required to add DiRe can be parallelized across multiple GPUs, depending upon availability. 
\begin{table}[hbt!]
\caption{Comparison of total time taken and GPU memory consumed for synthesizing IPC=$10$ images from ImageNet-1K dataset.}
\label{timing_analysis_result_ilsvrc}
\begin{center}
\begin{tabular}{l|c|c}
\toprule
Method & Time (in Hrs.) & Memory (in GB) \\
\midrule
SRe$^2$L & 2.94 & 10.6 \\
SRe$^2$L + DiRe & 3.13 & 13.0 \\
\bottomrule
\end{tabular}
\end{center}
\end{table}

\newpage
\section{Intra-class cosine similarity for ILSVRC}
\begin{figure}[h!]
\begin{center}
\centerline{\includegraphics[width=\columnwidth]{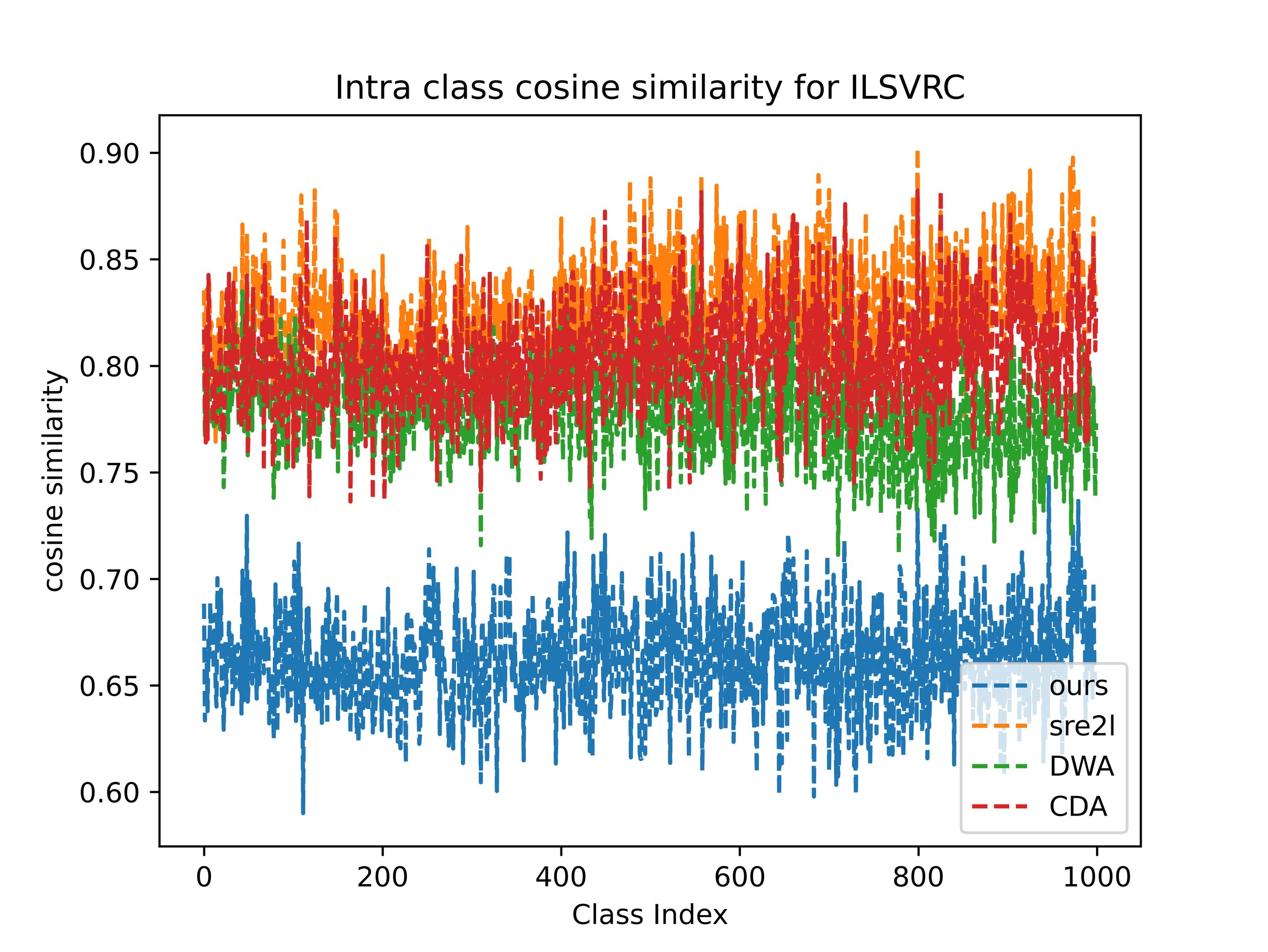}}
\caption{Intra-class cosine similarity of synthetic images generated from ImageNet-1K. We observe that DiRe lowers the cosine similarity across all classes, signifying improving diversity among the generated images. }
\label{ilsvrc_intra_class_cosine}
\end{center}
\end{figure}

Figure \ref{ilsvrc_intra_class_cosine} shows that, for all classes in ImageNet-1K, DiRe lowers intra-class cosine similarity, which is an indicator of improving diversity. 

\end{document}